\documentclass[11pt,letterpaper]{article}
\usepackage[T1]{fontenc}
\usepackage{xcolor}
\usepackage[letterpaper,textwidth=7.05in,textheight=9.35in,top=0.80in,headheight=14pt,headsep=11pt,footskip=23pt]{geometry}
\usepackage{fancyhdr,titlesec,tgheros,tcolorbox}
\usepackage{tikz}
\usepackage{newtxtext}
\usepackage{amsmath}
\usepackage{newtxmath}
\usepackage[hyphens]{url}
\usepackage{graphicx}
\usepackage[font=small,labelfont=bf,labelsep=period,hypcap=false]{caption}
\usepackage{booktabs}
\usepackage{multirow}
\usepackage{array}
\usepackage{placeins}
\usepackage{microtype}
\usepackage[numbers,sort&compress]{natbib}
\definecolor{paperInk}{HTML}{2D263B}
\definecolor{paperAccent}{HTML}{5E4B81}
\definecolor{paperPanel}{HTML}{EFEDF3}
\definecolor{paperRule}{HTML}{D3CBDF}
\definecolor{paperSecondary}{HTML}{574B68}
\usepackage[breaklinks,colorlinks,allcolors=paperAccent]{hyperref}
\hypersetup{pdftitle={EgoPathBench: Evaluating Zero-Shot Egocentric Waypoint Decision-Making in Vision-Language Models},pdfauthor={Yang Zhao, Zhuo Chen, Xubo Yang}}
\titleformat{\section}{\Large\sffamily\bfseries\color{paperInk}}{\thesection}{0.65em}{}
\titleformat{\subsection}{\large\sffamily\bfseries\color{paperInk}}{\thesubsection}{0.65em}{}
\titleformat{\subsubsection}{\normalsize\sffamily\bfseries\color{paperInk}}{\thesubsubsection}{0.65em}{}
\titlespacing*{\section}{0pt}{1.15em plus 2pt minus 2pt}{0.45em}
\titlespacing*{\subsection}{0pt}{0.9em plus 2pt minus 2pt}{0.35em}
\begin{document}
\thispagestyle{empty}
\enlargethispage{12pt}
\pdfbookmark[0]{Title and Abstract}{paper-cover}
\begin{tcolorbox}[colback=paperPanel,colframe=paperPanel,boxrule=0pt,
  sharp corners,left=17pt,right=17pt,top=14pt,bottom=12pt,
  boxsep=0pt,before skip=0pt,after skip=0pt]
\setlength{\parskip}{0pt}
{\sffamily\bfseries\color{paperInk}\fontsize{36}{39}\selectfont EgoPathBench\par}
\vspace{6pt}
{\sffamily\color{paperInk}\fontsize{16}{20}\selectfont
Evaluating Zero-Shot Egocentric Waypoint Decision-Making\\
in Vision-Language Models\par}
\vspace{10pt}
{\sffamily\bfseries Yang Zhao\quad Zhuo Chen\quad Xubo Yang\textsuperscript{*}\par}
\vspace{3pt}
{\small\sffamily\color{paperSecondary}Shanghai Jiao Tong University\quad
\textsuperscript{*}Correspondence: \href{mailto:yangxubo@sjtu.edu.cn}{yangxubo@sjtu.edu.cn}\par}
\vspace{7pt}
{\color{paperRule}\rule{\linewidth}{0.55pt}\par}
\vspace{7pt}
{\sffamily\bfseries\color{paperInk}Abstract\par}
\vspace{4pt}
{\small Widely used spatial-QA benchmarks increasingly motivate embodied navigation, yet evaluate foundation vision-language models (VLMs) through isolated queries about space rather than feasible action sequences in space. However, zero-shot waypoint navigation requires turning spatial understanding into a feasible action sequence: from a single first-person view, a model must jointly ground the target, respect agent-specific clearance, maintain edge-by-edge legality, and stop at the goal. To evaluate this capability, we introduce EgoPathBench, a dataset and five-task benchmark for first-person waypoint decision-making. Each example pairs an egocentric RGB image, a natural-language goal, and numbered visible waypoints, and scores predicted candidate sets or ordered routes against registered 3D scene geometry under task-specific Point or Embodied feasibility graphs. EgoPathBench comprises 31,852 training, 1,345 validation, and 1,111 benchmark questions, with at least one verified reference route for every route question. Across ten foundation VLMs, the highest EgoPath Score is only 35.6. While models often make locally plausible choices and the top model reaches 41.7\% success on Point Path, it drops to 10.0\% on both Embodied Path and Intent Path, with the remaining nine models below 3\% on Embodied Path (and at most 4.0\% on Intent Path)---exposing a sharp gap between local plausibility and complete embodied routes. Beyond evaluation, we construct the EgoPath-SFT training set. Fine-tuning Qwen3.5-4B on EgoPath-SFT raises its EgoPath Score from 3.9 to 38.9, improves closed-loop Isaac Sim execution by 20 percentage points, and transfers consistently to three external spatial benchmarks.
\par}
\vspace{9pt}
{\small\sffamily\bfseries
\href{https://egopathbench.github.io/EgoPathBench/}{Project page}
\quad\textcolor{paperSecondary}{\textbar}\quad
\href{https://huggingface.co/datasets/runder1/EgoPathBench}{Dataset}
\quad\textcolor{paperSecondary}{\textbar}\quad
\href{https://egopathbench.github.io/EgoPathBench/\#leaderboard}{Online leaderboard}\par}
\end{tcolorbox}

\vspace{10pt}
\noindent\begin{minipage}{\textwidth}
\centering
\begin{tikzpicture}[x=1pt,y=1pt,inner sep=0pt,font=\sffamily,
  heading/.style={font=\sffamily\bfseries,text=paperInk,anchor=west},
  model/.style={font=\sffamily\small,text=paperInk,anchor=east},
  tick/.style={font=\sffamily\footnotesize,text=paperSecondary},
  value/.style={font=\sffamily\footnotesize\bfseries,text=paperInk},
  grid/.style={draw=paperRule!66,line width=0.45pt}]
\path[use as bounding box] (0,13) rectangle (509,250);
\node[heading,font=\sffamily\large\bfseries] at (0,242) {EgoPathBench leaderboard};
\draw[paperRule,line width=0.55pt] (0,229) -- (509,229);

\node[heading] at (0,214) {Zero-shot model ranking};
\node[tick,anchor=west] at (0,198) {EgoPath Score};
\foreach \score in {0,10,20,30,40} {
  \draw[grid] ({99+3.5*\score},35) -- ({99+3.5*\score},190);
  \node[tick] at ({99+3.5*\score},25) {\score};
}
\foreach \name/\score/\barcolor [count=\rank from 0] in {
  GPT-6 Astra/35.6/paperAccent,
  Gemini 3.1 Pro/28.3/paperAccent!32,
  GPT-5.5/27.3/paperAccent!32,
  Claude Opus 4.8/25.6/paperAccent!32,
  MiniMax M3/21.8/paperAccent!32,
  Qwen 3.6/16.4/paperAccent!32,
  Mistral Large 3/15.8/paperAccent!32,
  Llama 4 Maverick/14.9/paperAccent!32,
  Kimi K2.6/9.7/paperAccent!32,
  Grok 4.3 Fast/1.4/paperAccent!32
}{
  \pgfmathsetmacro{\rowy}{181-15*\rank}
  \node[model] at (93,\rowy) {\name};
  \fill[\barcolor] (99,{\rowy-4}) rectangle ({99+3.5*\score},{\rowy+4});
  \node[value,anchor=west] at ({103+3.5*\score},\rowy) {\score};
}

\draw[paperRule,line width=0.55pt] (257,22) -- (257,219);
\node[heading] at (273,214) {Route success rate (\%)};
\foreach \start/\name/\barcolor in {
  274/GPT-6 Astra/paperAccent,
  351/Gemini 3.1 Pro/paperAccent!54,
  443/GPT-5.5/paperSecondary!23
}{
  \fill[\barcolor] (\start,193) rectangle ({\start+7},200);
  \node[tick,anchor=west] at ({\start+11},196.5) {\name};
}
\foreach \rate in {0,25,50} {
  \draw[grid] (291,{101+1.57*\rate}) -- (509,{101+1.57*\rate});
  \node[tick,anchor=east] at (287,{101+1.57*\rate}) {\rate};
}
\foreach \center/\task/\first/\second/\third in {
  324/Point/41.7/35.9/31.1,
  400/Embodied/10.0/2.9/1.3,
  476/Intent/10.0/4.0/1.5
}{
  \node[tick] at (\center,90) {\task};
  \foreach \offset/\rate/\barcolor in {
    -16/\first/paperAccent,
    0/\second/paperAccent!54,
    16/\third/paperSecondary!23
  }{
    \fill[\barcolor] ({\center+\offset-6.5},101)
      rectangle ({\center+\offset+6.5},{101+1.57*\rate});
    \node[value,anchor=south,font=\sffamily\fontsize{7.2}{9}\selectfont]
      at ({\center+\offset},{104+1.57*\rate}) {\rate};
  }
}

\fill[paperPanel,rounded corners=3pt] (273,20) rectangle (509,72);
\node[heading,font=\sffamily\small\bfseries] at (283,58) {Qwen3.5-4B fine-tuning};
\node[tick,anchor=west] at (283,35) {EgoPath Score};
\node[value,font=\sffamily\large\bfseries] at (380,43) {3.9};
\node[tick] at (380,29) {Base};
\draw[->,paperAccent,line width=0.7pt] (396,43) -- (439,43);
\node[value,font=\sffamily\large\bfseries,text=paperAccent] at (469,43) {38.9};
\node[tick] at (469,29) {+ EgoPath-SFT};
\end{tikzpicture}
\captionof{figure}{The ranking shows all ten zero-shot models. The three highest-ranked models are compared on route success, which requires reaching the goal along a legal route. Table~\ref{tab:main_leaderboard} gives the full results.}
\label{fig:cover_leaderboard}
\end{minipage}\par

\clearpage
\begin{figure*}[t]
\centering
\includegraphics[width=0.99\textwidth]{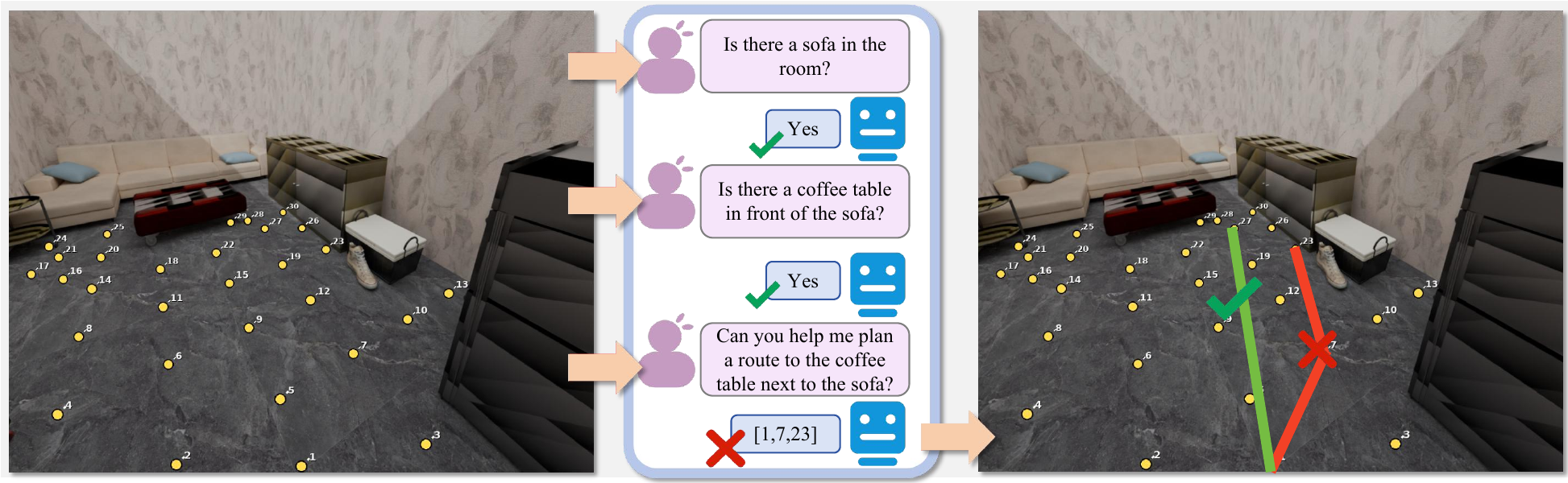}
\caption{Conceptual motivation for EgoPathBench. On the same first-person scene, familiar recognition and local spatial questions can be answered correctly. The right panel contrasts a complete, goal-reaching route with an incorrect route that violates the route and goal criteria. This contrast motivates evaluating integrated, embodiment-aware route decisions rather than component visual questions alone.}
\label{fig:overview}
\end{figure*}

\section{Introduction}

Foundation vision-language models are increasingly used in embodied navigation to translate first-person observations and task goals into spatial actions. Recent work has explored VLMs or LLMs for explicit navigation reasoning, generalist navigation, visual next-step planning, and zero-shot ObjectNav \cite{navgpt2024,navillm2024,navid2024,clcotnav2025}. Navigation decisions require more than recognizing objects or answering an individual spatial-relation question: a model must interpret the current environment, the intended target, and the available actions to make a goal-directed spatial choice. The integrated spatial intelligence of foundation VLMs is therefore an important basis for zero-shot navigation to generalize across new scenes and goals.

Existing evaluations leave a three-way gap when measuring a foundation VLM's own spatial decision-making for navigation. Spatial-intelligence benchmarks such as SpatialVLM, SpatialEval, VSI-Bench, 3DSRBench, EmbSpatial-Bench, EgoThink, and OpenEQA consistently motivate embodied navigation and robotics, yet evaluate models on isolated queries \emph{about} space---metric distances, pairwise relations, or symbolic turn choices---rather than feasible action sequences \emph{in} space \cite{spatialvlm2024,spatialeval2024,vsibench2025,threedsrbench2025,embspatial2024,egothink2024,openeqa2024}. Conversely, closed-loop 3D navigation benchmarks such as NavBench \cite{navbench2025} verify physical execution, but entangle single-view VLM reasoning with mapping, memory, low-level obstacle avoidance, and step-by-step replanning at high simulation cost. Closest to our setting, NaviTrace \cite{navitrace2025} isolates single-image route prediction, yet evaluates 2D image-space traces via similarity to human demonstrations and 2D pixel semantics rather than 3D scene geometry, making it unable to verify 3D embodied clearance or check vertex- and edge-level geometric legality under specified body constraints.

To bridge this gap, we introduce EgoPathBench, a dataset and five-task benchmark for first-person waypoint decision-making. Each example presents an egocentric RGB image with numbered visible waypoints and a natural-language prompt; the model returns a JSON list of traversable candidates or an ordered route. To test the progression from local perception to integrated embodied planning on a shared observation and action space, EgoPathBench organizes five tasks: \textbf{Point Traversability} and \textbf{Embodied Traversability} ask which visible waypoints are walkable versus traversable by a 0.6\,m-diameter robot; \textbf{Point Path} and \textbf{Embodied Path} share the same view, explicit target, and candidate set while evaluating complete routes on the Point and Embodied feasibility graphs; and \textbf{Intent Path} replaces the explicit target name with an intent and visual cue, requiring target grounding together with embodied route selection (Figure~\ref{fig:overview}).

To align visual observations with physical action consequences, every displayed waypoint, target region, and legal edge is anchored to registered 3D scene geometry prior to language generation. Any predicted route is scored directly by candidate feasibility, required start, adjacent-edge legality, acceptable goal arrival, and route efficiency on the corresponding Point or Embodied graph---without requiring imitation of a single reference trajectory. Every route question retains at least one scene-verified reference route visible in the current view. Across splits, EgoPathBench provides 31,852 training (EgoPath-SFT), 1,345 validation, and 1,111 benchmark questions.

Evaluating ten foundation VLMs on EgoPathBench exposes a stark gap between local plausibility and complete embodied routes. While models frequently pick a legal first step and the top-ranked model (GPT-6 Astra) matches human accuracy on local traversability and reaches 41.7\% success on Point Path (with a top EgoPath Score of 35.6), on the exact same views and targets, its success drops to 10.0\% on Embodied Path and Intent Path, while the other nine fall below 3\% on Embodied Path (at most 4.0\% on Intent Path). Isolated follow-up queries further show that even when GPT-6 Astra correctly judges the individual edges of its own predicted route, most of its full plans still fail---failing to maintain multi-step clearance or stop at the goal. Finally, fine-tuning Qwen3.5-4B on EgoPath-SFT raises its EgoPath Score from 3.9 to 38.9, improves closed-loop Isaac Sim execution by 20 percentage points, and yields consistent gains across three external spatial benchmarks.

In summary, our contributions are three-fold:
\begin{itemize}
    \item We formulate EgoPathBench, a 3D scene-grounded benchmark with five paired tasks that bridges isolated local spatial judgments and integrated first-person waypoint planning under explicit Point and Embodied feasibility rules.
    \item We construct and release 31,852 training (EgoPath-SFT), 1,345 validation, and 1,111 benchmark questions by aligning rendered views, visible waypoints, and resolved targets with 3D feasibility graphs.
    \item We evaluate ten foundation VLMs, diagnose why locally plausible predictions collapse on embodied routes, and validate EgoPath-SFT on EgoPathBench, closed-loop Isaac Sim rollouts, and three external spatial benchmarks.
\end{itemize}

\section{Related Work}

\paragraph{Foundation-model navigation and evaluation.}
R2R, REVERIE, RxR, and VLN-CE established vision-language navigation in discrete and continuous indoor environments \cite{r2r2018,reverie2020,rxr2020,vlnce2020}. Foundation-model agents such as NavGPT, NaviLLM, and NaVid subsequently used language reasoning, generalist embodied modeling, and video-based planning for navigation \cite{navgpt2024,navillm2024,navid2024}. Building on visual-prompting schemes that overlay numbered markers on RGB images \cite{yang2023setofmark,nasiriany2024pivot,goetting2025vlmnav}, recent zero-shot systems organize spatial decisions through different interfaces: SmartWay combines waypoint prediction with history-aware backtracking; VLFM, InstructNav, and CA-Nav use occupancy or value maps and sub-instruction constraints; AgenticNav and P2DNav expose pixel-level or hierarchical direction-to-grounding actions; and DreamNav predicts trajectories rather than isolated points \cite{smartway2025,vlfm2024,instructnav2024,canav2025,agenticnav2026,p2dnav2026,dreamnav2025}. Open-Nav studies spatio-temporal reasoning with open-source models, Nav-R1 learns structured navigation traces, and LHPR-VLN extends evaluation to decision consistency across long-horizon subtasks \cite{opennav2025,navr12025,lhprvln2025}. Across full navigation systems, outcomes can depend on different combinations of candidate generation, mapping, memory, control, and replanning over repeated observations. EgoPathBench measures a controlled capability within this process: given the current first-person observation and a shared vocabulary of visible spatial actions, can a foundation VLM form a geometrically feasible and goal-consistent waypoint decision?

NaviTrace is the closest evaluation to our setting. It presents a single real RGB image, a navigation instruction, and an embodiment description, and asks a VLM to produce a continuous trace in image space \cite{navitrace2025}. The prediction is scored by Dynamic Time Warping against human-annotated traces, endpoint error, and embodiment-conditioned penalties derived from pixel semantics. NaviTrace therefore measures agreement with expert navigation demonstrations under a score that combines trace similarity, endpoint accuracy, and embodiment-conditioned semantic penalties.

EgoPathBench differs in the supervision associated with each observation. Each image is registered to an underlying 3D scene representation, from which we derive visible waypoints, agent-specific feasibility, direct traversability between waypoints, and acceptable goal regions. This scene backing allows an arbitrary predicted route to be evaluated by the geometric outcomes of its selected actions: whether its waypoints are feasible, whether every consecutive transition satisfies the specified agent constraints, and whether the route reaches the goal. A reference route certifies that a geometrically valid solution exists, but is not the unique trajectory that a prediction must imitate. Thus, NaviTrace evaluates trajectory agreement defined by expert demonstrations and image semantics, whereas EgoPathBench evaluates waypoint-action outcomes defined by scene geometry.

\paragraph{Spatial intelligence and embodied reasoning.}
Existing spatial-intelligence benchmarks examine complementary components of foundation VLMs' spatial ability. SpatialVLM and SpatialEval emphasize distance, direction, and spatial relations; VSI-Bench studies video-based spatial understanding and memory; ViewSpatial evaluates multi-perspective spatial localization; and 3DSRBench focuses on 3D structure \cite{spatialvlm2024,spatialeval2024,vsibench2025,viewspatial2025,threedsrbench2025}. EmbSpatial-Bench, EgoThink, and OpenEQA introduce first-person or embodied observations \cite{embspatial2024,egothink2024,openeqa2024}. Embodied3DBench further evaluates low-level embodied skills including grounding, affordance, and trajectory prediction; CapNav studies capability-conditioned navigation under agent-specific mobility constraints; and IndustryNav evaluates active planning and collision-aware navigation in dynamic industrial environments \cite{embodied3dbench2026,capnav2026,industrynav2025}. Together, these benchmarks show that embodied spatial intelligence involves not only recognizing scene relations, but also determining where an agent can act and what spatial consequences its actions produce.

EgoPathBench organizes these abilities into a joint decision with an explicit action interpretation. A model must relate the target, candidate locations, agent constraints, and route structure in the current observation, then return either a traversable waypoint set or an ordered route. The prediction is evaluated through geometry-defined candidate feasibility, consecutive-edge legality, and goal attainment. EgoPathBench thereby connects low-level embodied spatial cues to ordered first-person waypoint decisions with explicit scene-grounded consequences.

\section{EgoPathBench Dataset and Benchmark}

\begin{figure*}[t]
\centering
\includegraphics[width=0.99\textwidth]{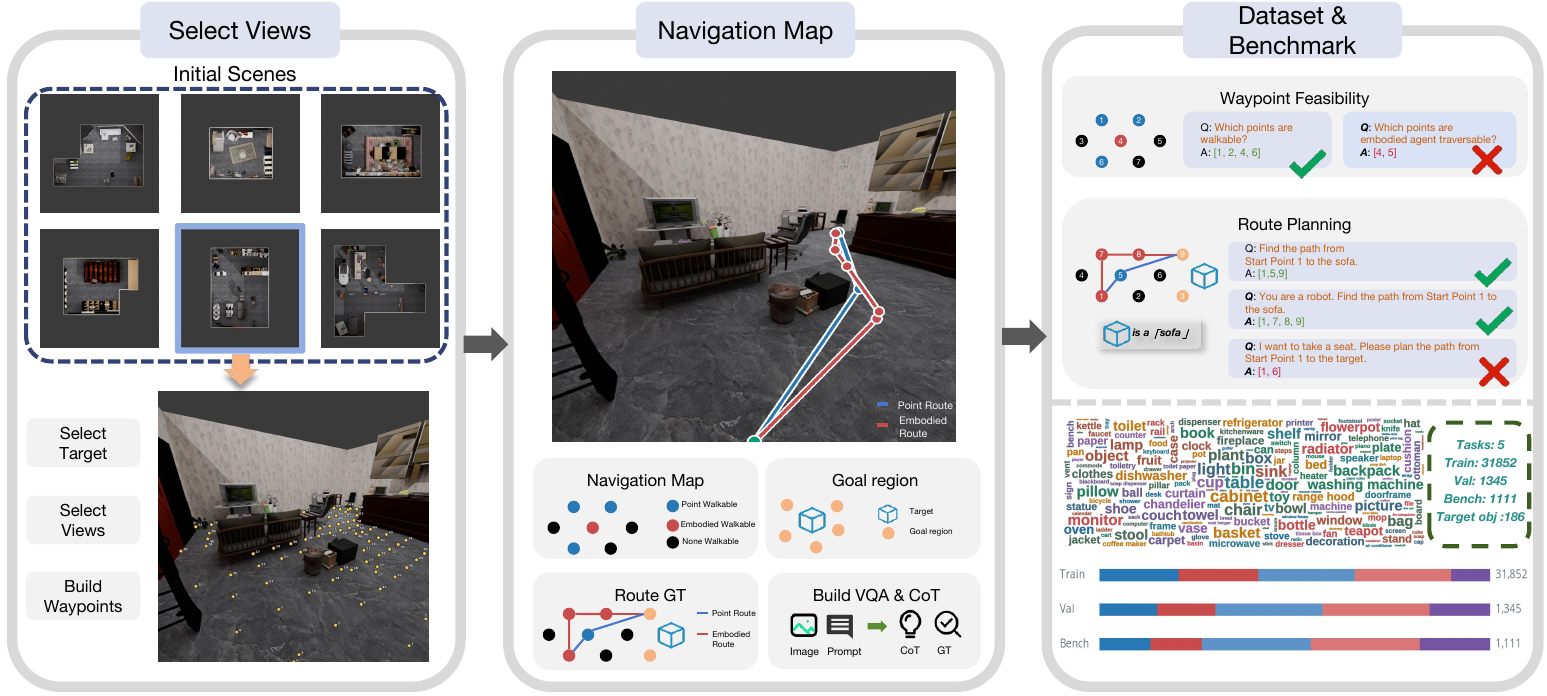}
\caption{EgoPathBench construction pipeline. We select scenes, first-person views, and targets, then build visible waypoints, point- and embodiment-specific navigation maps, goal regions, and geometry-verified reference routes. These fixed scene annotations support image-question and Spatial CoT generation, yielding two waypoint-feasibility tasks and three route-planning tasks across the training, validation, and benchmark splits.}
\label{fig:construction_pipeline}
\end{figure*}

\subsection{Benchmark Formulation}

EgoPathBench represents navigation decisions as selections over a visible waypoint vocabulary, mirroring the visual action interface used by recent zero-shot VLM navigation systems \cite{smartway2025,agenticnav2026,p2dnav2026}. Each example provides an egocentric RGB image with numbered candidate locations and a natural-language task. The model returns a JSON list of display IDs. For the two traversability tasks, the list denotes an unordered set of traversable candidates. For the three path tasks, it denotes an ordered route from a specified start to an acceptable goal.

Formally, an example consists of an egocentric observation $o$, a displayed waypoint set $A$, a target specification $t$, and a task-specific feasibility graph $G$. Every display ID in $A$ is anchored to a 3D scene location. Given $o$, $A$, and $t$, the model predicts either a subset of $A$ or an ordered sequence over $A$. Traversability tasks evaluate the feasibility of selected vertices. Route tasks additionally require the specified start, legal consecutive edges in $G$, and an endpoint in the acceptable goal set.

Point and Embodied tasks share the same observation, target object, and visible waypoint vocabulary, while differing in their task-specific feasibility graphs, agent-specific prompt specifications, and acceptable goal regions (Appendix~\ref{app:evaluation_protocol}). Specifically, feasibility graphs are built on a height-filtered 2D occupancy grid (inflated by a 0.30\,m radius for Embodied/Intent; details in Appendix~\ref{app:evaluation_protocol}), where an edge is legal if its straight-line swept corridor is collision-free and acceptable endpoints are defined near the target box ($d_{\min}+0.10$\,m, with Point-task extension detailed in Appendix~\ref{app:evaluation_protocol}). Predictions are evaluated on the corresponding Point or Embodied graph. The Point graph is constructed without footprint expansion, while the Embodied graph accounts for the 0.6\,m robot diameter specified in the task prompt. The same visually selected action can therefore have different consequences under the two evaluation rules.

Thus, the model acts solely from the current observation, while the consequences of its choices are evaluated against registered 3D scene geometry.

\subsection{Five-Task Design}

EgoPathBench contains two traversability tasks and three route tasks. The five tasks progressively introduce target grounding, embodiment, and route-composition requirements while retaining a common input and output interface. Table~\ref{tab:task_taxonomy} in Appendix~\ref{app:evaluation_protocol} summarizes the five tasks.

\textbf{Point Traversability} asks the model to select all walkable candidates. \textbf{Embodied Traversability} uses the same scene view and candidate space and specifies a robot diameter of 0.6\,m, requiring sufficient clearance at the selected locations. This paired design tests how embodiment changes action feasibility within the same observation.

\textbf{Point Path} provides an explicit target and asks for an ordered route to it. \textbf{Embodied Path} holds the target, view, and candidate space fixed, adds the robot dimensions to the prompt, and evaluates the route on the Embodied graph. \textbf{Intent Path} replaces the explicit object name with an intent and a visual cue. It therefore requires the model to resolve the intended object before selecting an embodied route.

\subsection{Data Construction and Annotation}

Figure~\ref{fig:construction_pipeline} summarizes the construction process. We first establish scene provenance and first-person scene--view records, then generate visible waypoints and geometric task labels. Natural-language questions and Spatial CoT are produced only after the target, action space, and reference route have been fixed. Quality filtering and challenge-set selection produce the final release.

\noindent\textbf{Scenes, views, and targets.}
We use unified simulatable indoor assets normalized through InternScenes \cite{internscenes2025}. The scenes retained in our release derive from 3RScan \cite{threerscan2019}, ScanNet \cite{scannet2017}, ARKitScenes \cite{arkitscenes2021}, and Matterport3D \cite{matterport3d2017}. We use these assets for scene geometry and rendering, and construct our own first-person observations, waypoint annotations, route labels, and question data. For each scene, we sample first-person cameras in traversable space and remove views whose immediate camera neighborhood is obstructed. Target instances are then filtered by frustum projection, observation distance, projected size, and visible surface evidence, excluding targets that are outside the image, too small, or heavily occluded. Shared camera parameters and scene geometry bind each retained target and RGB observation to a common scene--view record, which provides the spatial basis for subsequent waypoint and route annotation.

\noindent\textbf{Waypoints and geometric labels.}
For each retained view, we project two types of markers into the image: ground action candidates and selected visible object- or structural-surface locations used as non-traversable negatives. Depth, ray-visibility, and marker-spacing checks associate every displayed ID with a distinct scene location while reducing marker overlap and foreground occlusion. Route tasks use visible ground waypoints and explicitly insert the waypoints required by retained reference routes. We evaluate ground actions under both point and embodied geometry. Traversability examples therefore mix feasible ground actions, clearance-sensitive ground actions, and visible surface negatives, so the task cannot be solved by selecting every displayed ID. Point and Embodied Traversability share the same view, while their labels follow the corresponding agent-feasibility conditions. Each displayed candidate stores its ID, image projection, and scene coordinate; the numbered RGB overlay and the geometric annotations therefore refer to the same action vocabulary.

\noindent\textbf{Paired route construction.}
For each route unit, we anchor the start on visible floor near the bottom of the image and generate feasible goal locations around the target footprint. We then search separately through point and embodied free space, convert each continuous path into sparse route waypoints, and project the required waypoints back into the current image. Construction checks route connectivity, task-specific feasibility, arrival in the target region, and the availability of every required waypoint through the displayed action interface. Routes that fail these checks are discarded. Point Path and Embodied Path consequently share the target, start, and scene view while retaining routes validated for their respective agent geometries. Every retained route question has at least one scene-verified reference route to its target.

\noindent\textbf{Question text and Spatial CoT.}
Target identity, displayed waypoints, goal region, and reference-route identity are fixed before language generation. Explicit-target tasks name the required object directly. For Intent Path, we construct a view-specific candidate universe from visible objects and generate a target description using supported relations, attributes, colors, or distance cues. The resulting description is resolved again against the objects available in the current view. We retain it only when it uniquely identifies the fixed target without unsupported cues. Intent Path is derived from an accepted Embodied Path instance, so changing the linguistic specification does not change its target or route geometry. Each training example in EgoPath-SFT pairs a Spatial CoT rationale with its geometry-verified final answer. GPT-5.5 verbalizes the fixed target, candidates, feasibility labels, legal edges, and reference route into task-specific reasoning text. The exported answer is checked against the formal annotation, keeping language generation downstream of the geometric ground truth.

\subsection{Quality Control}

Quality control applies geometry and view--action checks before language is attached. Candidate locations, route edges, and target regions share one scene coordinate system; disconnected or colliding routes, routes that miss the target region, and examples without required display waypoints are removed. The RGB image, waypoint overlay, and ID-to-waypoint record come from the same rendered view, and point and embodied routes are validated under their corresponding agent geometries. Prompts may describe only targets and scene facts fixed during construction. Explicit-target questions are checked against the selected target, while intent questions must resolve uniquely among objects visible from the current view. A final cross-check verifies consistency across targets, prompts, displayed waypoints, reference routes, and answers. Ambiguous or unsupported prompts are repaired or removed without changing the established geometric labels or route identities.

\subsection{Benchmark Selection and Release}

After constructing the full question pool, we select the formal benchmark as a challenging subset emphasizing scene clutter, traversability boundaries, point--embodied feasibility differences, competing targets, and route composition. Selection operates on complete task bundles. Point and Embodied Traversability form a paired view bundle; Point Path and Embodied Path form a paired route bundle with a shared target and candidate space; eligible Intent Path questions accompany their corresponding embodied routes. Bundle-level selection prevents the benchmark from retaining only one side of a paired comparison. We then select scenes while balancing task coverage, geometric difficulty, and scene diversity and limiting concentration within InternScenes. Splits are formed by source group, so the same original scan or related regions do not cross training, validation, and benchmark sets. The final release contains 31,852 training questions, 1,345 validation questions, and 1,111 benchmark questions across single-room indoor scenes without stairs, and all annotations, prompts, and EgoPath-SFT data will be released for research use.

\section{Experiments}

\begin{table*}[t]
\centering
{
\scriptsize
\setlength{\tabcolsep}{2.4pt}
\resizebox{\textwidth}{!}{%
\begin{tabular}{@{}lcccccccccccccc@{}}
\toprule
& & \multicolumn{2}{c}{Point Trav.} & \multicolumn{2}{c}{Embodied Trav.} & \multicolumn{3}{c}{Point Path} & \multicolumn{3}{c}{Embodied Path} & \multicolumn{3}{c}{Intent Path} \\
\cmidrule(lr){3-4}\cmidrule(lr){5-6}\cmidrule(lr){7-9}\cmidrule(lr){10-12}\cmidrule(lr){13-15}
Model & Score & BA & F1 & BA & F1 & VPR & SR & SPL & VPR & SR & SPL & VPR & SR & SPL \\
\midrule
\multicolumn{15}{@{}l}{\emph{Zero-shot foundation VLMs}} \\
GPT-6 Astra & \textbf{35.6} & \textbf{83.1} & \textbf{83.0} & 75.1 & \textbf{63.6} & \textbf{72.5} & \textbf{41.7} & \textbf{35.7} & \textbf{41.1} & \textbf{10.0} & \textbf{9.5} & \textbf{35.8} & \textbf{10.0} & \textbf{9.3} \\
Gemini 3.1 Pro & 28.3 & 76.5 & 79.0 & 72.7 & 56.2 & 63.7 & 35.9 & 28.7 & 9.1 & 2.9 & 2.4 & 10.4 & 4.0 & 3.3 \\
GPT-5.5 & 27.3 & 74.1 & 77.1 & \textbf{77.3} & 62.5 & 60.5 & 31.1 & 24.8 & 5.5 & 1.3 & 1.3 & 9.0 & 1.5 & 1.4 \\
Claude Opus 4.8 & 25.6 & 77.9 & 74.3 & 73.5 & 59.6 & 66.0 & 21.0 & 16.7 & 12.0 & 1.6 & 1.5 & 14.4 & 2.5 & 2.2 \\
MiniMax M3 & 21.8 & 77.0 & 76.2 & 68.7 & 52.8 & 50.8 & 13.3 & 8.7 & 5.5 & 1.9 & 1.7 & 9.0 & 2.5 & 2.3 \\
Qwen 3.6 & 16.4 & 67.4 & 72.1 & 64.8 & 49.1 & 49.2 & 15.2 & 10.9 & 2.6 & 1.0 & 0.9 & 5.0 & 1.5 & 1.3 \\
Mistral Large 3 & 15.8 & 65.2 & 70.8 & 68.5 & 52.5 & 35.6 & 11.0 & 5.8 & 0.3 & 0.0 & 0.0 & 3.5 & 0.5 & 0.5 \\
Llama 4 Maverick & 14.9 & 66.8 & 66.5 & 66.0 & 49.8 & 36.2 & 8.7 & 5.9 & 1.9 & 0.0 & 0.0 & 2.0 & 0.0 & 0.0 \\
Kimi K2.6 & 9.7 & 60.2 & 69.8 & 57.8 & 44.2 & 40.5 & 11.7 & 8.2 & 1.3 & 0.7 & 0.5 & 1.5 & 0.5 & 0.4 \\
Grok 4.3 Fast & 1.4 & 52.3 & 58.4 & 50.0 & 35.8 & 18.4 & 2.6 & 1.2 & 5.2 & 0.0 & 0.0 & 3.5 & 0.0 & 0.0 \\
\midrule
\multicolumn{15}{@{}l}{\emph{Training-resource study}} \\
Qwen3.5-4B base & 3.9 & 54.6 & 55.6 & 54.9 & 33.1 & 9.1 & 0.7 & 0.1 & 0.3 & 0.0 & 0.0 & 0.0 & 0.0 & 0.0 \\
+ EgoPath-SFT & 38.9 & 89.3 & 89.2 & 83.4 & 71.2 & 77.0 & 31.4 & 28.6 & 34.9 & 7.1 & 6.9 & 44.8 & 10.4 & 10.0 \\
\bottomrule
\end{tabular}
}
}
\caption{EgoPathBench leaderboard (\%). The upper block compares zero-shot foundation VLMs; the lower block compares Qwen3.5-4B before and after fine-tuning on EgoPath-SFT. Score is the equal-weight five-task macro-average. BA/F1 evaluate traversability, while VPR, SR, and SPL evaluate route validity, goal-reaching success, and efficiency.}
\label{tab:main_leaderboard}
\end{table*}

\subsection{Experimental Setup}

The benchmark contains 1,111 questions: 146 Point Traversability, 146 Embodied Traversability, 309 Point Path, 309 Embodied Path, and 201 Intent Path. Traversability uses candidate-level balanced accuracy (BA), the mean recall across the task-feasible and infeasible classes, and F1, the precision--recall harmonic mean for feasible candidates. For route tasks, valid path rate (VPR) measures complete route legality, success rate (SR) additionally requires an acceptable endpoint, and success weighted by path length (SPL) discounts successful routes that are longer than the shortest legal reference while assigning zero to failures. EgoPath Score equally averages chance-adjusted traversability BA and route SR across the five tasks. Formal definitions are provided in the supplementary material.

We evaluate ten foundation VLMs. Every model receives the same image, task prompt, visible waypoint IDs, and JSON output protocol, and every prediction is scored by the same evaluator. Endpoint identifiers, generation settings, training hyperparameters, and control-specific protocols are provided in the supplementary material.

\subsection{Zero-Shot VLM Results}

Table~\ref{tab:main_leaderboard} gives the zero-shot leaderboard. GPT-6 Astra leads with an EgoPath Score of 35.6, followed by Gemini 3.1 Pro (28.3) and GPT-5.5 (27.3).

GPT-6 Astra achieves the best zero-shot Point Traversability BA (83.1\%) and F1 on both traversability tasks (83.0\% and 63.6\%). GPT-5.5 leads Embodied Traversability BA at 77.3\%. Astra leads all route metrics. On the paired Point and Embodied Path questions, the scene view, target, and waypoint set are shared. Astra's success rate drops from 41.7\% on Point Path to 10.0\% on Embodied Path. It also achieves 10.0\% success on Intent Path. The respective valid path rates are 72.5\%, 41.1\%, and 35.8\%. The gaps between validity and success reflect difficulty reaching an acceptable endpoint along a legal route. Rescoring all ten models under a shared 0.6\,m absolute goal radius without Point endpoint extension preserves this gap (24.14\% vs.\ 2.78\% pooled SR; Appendix~\ref{app:endpoint_controls}).

\begin{figure}[t]
\centering
\includegraphics[width=0.86\linewidth]{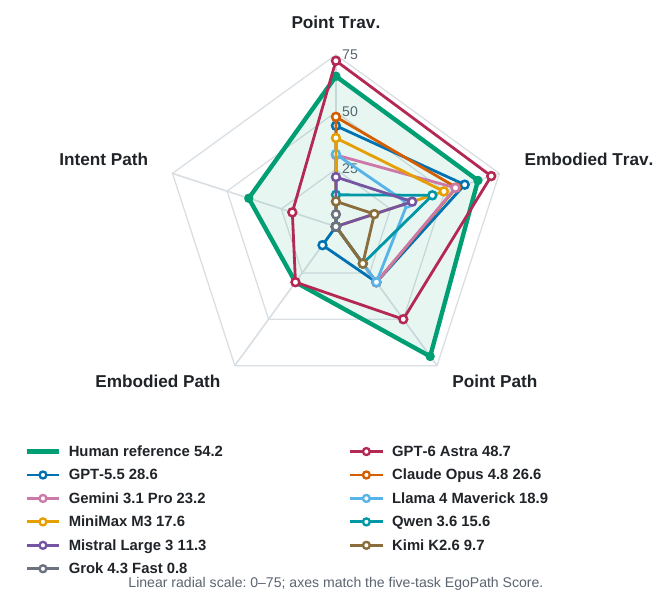}
\caption{Human reference versus ten VLMs on the same questions. Axes show $2\mathrm{BA}-1$ for traversability and route success rates; the legend reports EgoPath Score.}
\label{fig:human_calibration}
\end{figure}

Figure~\ref{fig:human_calibration} compares the ten models with a human reference on the same 50 questions, ten per task. Astra scores 48.7 versus 54.2 for the human reference. Astra achieves 86.2\% and 85.6\% BA on Point and Embodied Traversability, compared with 82.8\% and 82.6\% for the human reference. The human reference succeeds on 70\% of Point Path and 40\% of Intent Path questions, compared with 50\% and 20\% for Astra. Both achieve 30\% on Embodied Path. The rankings differ between local waypoint judgments and complete route planning. Appendix~\ref{app:human_examples} shows the corresponding qualitative comparisons.

\subsection{Single-Image Geometric Answerability}
\label{sec:geometric_answerability}

\begin{figure}[!ht]
\centering
\includegraphics[width=\linewidth]{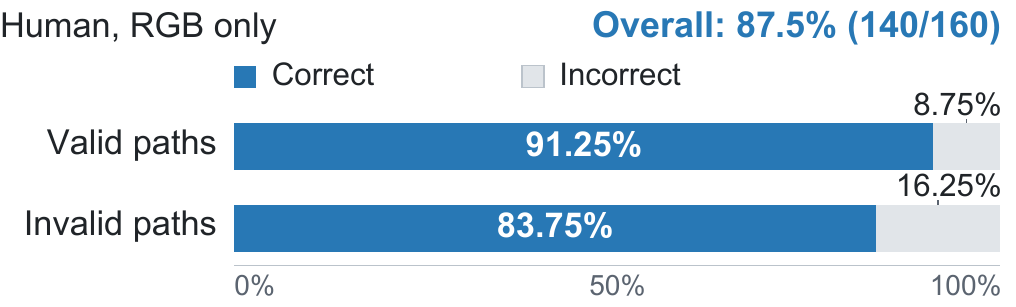}
\caption{RGB-only human assessment of geometric path validity by four participants (40 judgments each).}
\label{fig:rgb_human_validity}
\end{figure}

Given the sharp performance drop on embodied routes, we first verify whether path feasibility can be judged from a single egocentric RGB image. Because judging 0.6\,m clearance in single-view RGB is visually more ambiguous than point-mass feasibility on the same scenes---and Embodied Path shares identical traversability rules with Intent Path when goal resolution is excluded---we evaluate 40 model-predicted Embodied Path subpaths (20 valid and 20 invalid, spanning 2--4 waypoints). Four participants independently judge whether each of the 40 paths is traversable using only the original egocentric RGB image, numbered waypoints, the candidate path, and the specified embodiment constraints, without access to depth, 3D scene geometry, or validity labels. Participants P01--P04 correctly judge 39, 32, 35, and 34 paths, respectively, totaling 140 correct judgments out of 160. As shown in Figure~\ref{fig:rgb_human_validity}, human judgments achieve \textbf{91.25\%} accuracy on valid paths and \textbf{83.75\%} on invalid paths, with \textbf{87.5\%} overall accuracy. Appendix~\ref{app:geometric_validation} provides supporting geometric validation, visual-input controls, and human-study consent details.

\subsection{Where Complete Routes Fail}
\label{sec:route_diagnostics}

\begin{figure}[t]
\centering
\includegraphics[width=\linewidth]{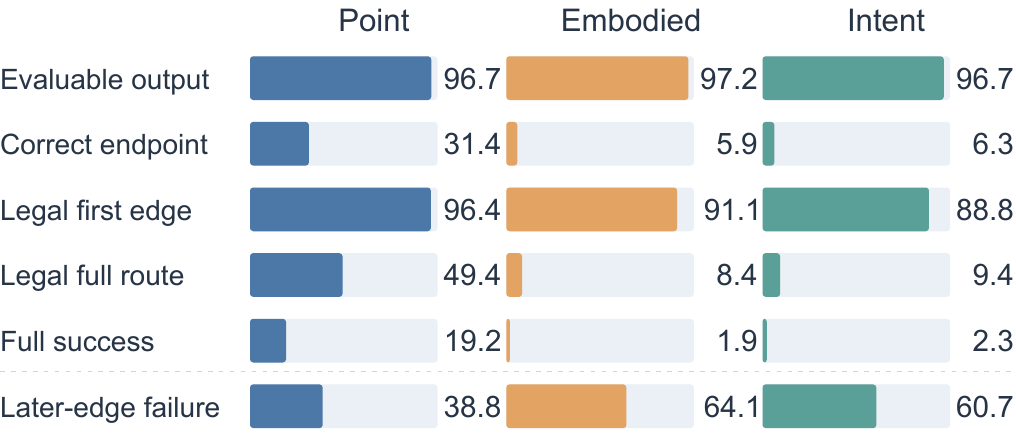}
\caption{Route diagnostics pooled over ten VLMs (\%). The first five rows are parallel rates over all predictions. Later-edge failure is conditioned on a legal first edge and an acceptable endpoint: 376/970, 107/167, and 71/117 for Point, Embodied, and Intent. Each bar spans 0--100\%.}
\label{fig:failure_decomposition}
\end{figure}

Since single-view path feasibility is human-verifiable (87.5\%), why do foundation VLMs still fail on complete routes? Figure~\ref{fig:failure_decomposition} pools the ten models. Evaluable outputs account for 96.7--97.2\% of predictions. Correct endpoint rates are 31.4\%, 5.9\%, and 6.3\% for Point, Embodied, and Intent Path, respectively.

Legal first-edge rates are 96.4\%, 91.1\%, and 88.8\%. Among predictions with a legal first edge, 48.8\%, 90.7\%, and 89.4\% contain an illegal later edge. Full-route legality is 49.4\%, 8.4\%, and 9.4\%, and joint success is 19.2\%, 1.9\%, and 2.3\%. The diagnostics distinguish endpoint selection and continued edge legality as separate challenges.

Among routes with a legal first edge and an acceptable endpoint, 38.8\%, 64.1\%, and 60.7\% contain an illegal later edge in Point, Embodied, and Intent Path, respectively. Rates on the fewer-waypoint subset are 39.3\%, 65.4\%, and 59.7\% (Table~\ref{tab:route_construct_bridge} in Appendix~\ref{app:dataset_stats}). In Embodied and Intent Path, illegal-edge rates rise from 86.3\% and 85.5\% on questions with one reference edge to 92.4\% and 90.2\% on questions with at least three reference edges (Table~\ref{tab:route_length_stratification} in Appendix~\ref{app:dataset_stats}).

\paragraph{Paired local judgments.}
To test whether route failures reflect missing local perception or a breakdown in composing full plans, we reuse GPT-6 Astra's plans for 50 questions across 25 scenes (20 Point, 20 Embodied, 10 Intent). Three independent probes use the same image and task context under identical high-effort reasoning settings to select an acceptable endpoint, list traversable candidates, and judge the legality of the plan's distinct directed edges in shuffled order. Table~\ref{tab:astra_local_probes} reports planning success rate (SR) and local-probe scores by task. Goal accuracy measures membership in the acceptable endpoint set. Point-set exact match (EM) requires the exact set of all feasible candidates. Edge-label EM requires every queried edge label to be correct.

\begin{table}[!ht]
\centering
{
\small
\setlength{\tabcolsep}{3pt}
\begin{tabular}{@{}lrcccc@{}}
\toprule
Task & $N$ & \shortstack{Planning\\SR} & \shortstack{Goal\\Acc.} & \shortstack{Point-set\\EM} & \shortstack{Edge-label\\EM} \\
\midrule
Point & 20 & 35.0 & 70.0 & 10.0 & 65.0 \\
Embodied & 20 & 0.0 & 5.0 & 0.0 & 30.0 \\
Intent & 10 & 0.0 & 20.0 & 0.0 & 40.0 \\
\midrule
All & 50 & 14.0 & 34.0 & 4.0 & 46.0 \\
\bottomrule
\end{tabular}
}
\caption{GPT-6 Astra planning and local-probe results by route task (\%). $N$ is the number of questions. Exact match (EM) requires the complete probe answer to match the annotation.}
\label{tab:astra_local_probes}
\end{table}

The results reveal three distinct bottlenecks. First, Astra shows an asymmetric edge bias: it identifies 308/318 (96.9\%) legal edges but only 21/59 (35.6\%) illegal edges, indicating that it misses nearly two-thirds of clearance or obstacle violations even in isolation; meanwhile, enumerating the full traversable set succeeds on only 2/50 questions (0\% on Embodied and Intent). Second, local knowledge does not reliably carry over to full-plan synthesis: among the 23/50 questions where Astra labels every edge of its plan correctly in isolation, 18 original plans still fail---including eight that traverse an edge Astra itself flags as illegal when queried separately, and ten that remain legal but miss the goal. Even on the single question with all three probes correct, the original plan fails. Third, adding an explicit stopping instruction to terminate at the nearest feasible goal waypoint raises goal hit rate from 26.0\% to 40.0\% and overall SR from 14.0\% to 24.0\% (Table~\ref{tab:stopping_instruction}), showing that explicit stopping guidance improves goal arrival for Astra, yet 76.0\% of its routes (and 90.0\% on Embodied and Intent Path) still fail under this prompt, indicating persistent difficulty in enforcing 3D embodied clearance across multi-step planning.

\begin{table}[!t]
\centering
\small
\setlength{\tabcolsep}{4pt}
\begin{tabular}{@{}lrcccc@{}}
\toprule
& & \multicolumn{2}{c}{Original} & \multicolumn{2}{c}{Explicit stopping} \\
\cmidrule(lr){3-4}\cmidrule(lr){5-6}
Task & $N$ & Goal Hit & SR & Goal Hit & SR \\
\midrule
Point & 20 & 55.0 & 35.0 & 60.0 & 45.0 \\
Embodied & 20 & 10.0 & 0.0 & 25.0 & 10.0 \\
Intent & 10 & 0.0 & 0.0 & 30.0 & 10.0 \\
All & 50 & 26.0 & 14.0 & 40.0 & 24.0 \\
\bottomrule
\end{tabular}
\caption{GPT-6 Astra stopping-instruction control (\%). Goal Hit is the fraction of planned routes whose final waypoint falls in the acceptable goal set; SR requires a legal route and an acceptable endpoint.}
\label{tab:stopping_instruction}
\end{table}

\FloatBarrier

\subsection{Training-Resource Evaluation}
\label{sec:training_resource_eval}

We fine-tune Qwen3.5-4B on EgoPath-SFT (the 31,852 training examples pairing Spatial CoT rationales with verified geometric answers). The base and fine-tuned (3,000-step; Appendix~\ref{app:evaluation_protocol}) models share the same EgoPathBench contract and items from VSI-Bench Route Planning, SpatialEval-VTQA, and 3DSRBench \cite{vsibench2025,spatialeval2024,threedsrbench2025}.

Table~\ref{tab:main_leaderboard} (bottom) shows the in-domain gain on the scene-disjoint split: EgoPath-SFT improves every metric to an EgoPath Score of 38.9, driven by traversability and Intent Path while zero-shot Astra still leads Point and Embodied Path SR.

\begin{table}[!ht]
\centering
{
\footnotesize
\setlength{\tabcolsep}{3pt}
\begin{tabular}{@{}llrrr@{}}
\toprule
Benchmark & Setting & Base & SFT & $\Delta$ \\
\midrule
\multirow{2}{*}{VSI-Bench Route Planning}
& Full & 29.38 & 33.51 & +4.13 \\
& Debiased & 20.18 & 24.56 & +4.38 \\
SpatialEval-VTQA & Full & 61.8 & 71.4 & +9.6 \\
3DSRBench & Full & 58.0 & 59.4 & +1.4 \\
\bottomrule
\end{tabular}
}
\caption{Performance of Qwen3.5-4B on three spatial benchmarks before and after supervised fine-tuning on EgoPath-SFT (\%).}
\label{tab:training_resource}
\end{table}

As shown in Table~\ref{tab:training_resource}, it also improves all external evaluations: +4.13 (Full) and +4.38 (Debiased) on VSI-Bench Route Planning, +9.6 on SpatialEval-VTQA, and +1.4 on 3DSRBench, confirming transfer across distinct multiple-choice and QA formats.

\paragraph{Closed-loop simulation.}
To test whether single-view gains transfer to multi-step execution, we evaluate both Qwen3.5-4B models on 50 fixed tasks across 25 held-out scenes (pre-verified by a reference policy). Following standard waypoint-based navigation \cite{smartway2025,p2dnav2026}, both models share the same egocentric waypoint-projection front-end and Isaac Sim controller. At each step, the model sees a fresh RGB view with numbered waypoints, and the controller executes up to 0.6\,m of the first segment. Success requires supported, contact-free arrival within 0.6\,m of the target's ground-projected box in 12 decisions.

\begin{table}[!ht]
\centering
{
\small
\setlength{\tabcolsep}{3pt}
\begin{tabular}{@{}lcc@{}}
\toprule
Model & Successes (SR) & \shortstack{Mean distance\\reduction (m)} \\
\midrule
Qwen3.5-4B base & 0/50 (0\%) & 0.452 \\
+ EgoPath-SFT & 10/50 (20\%) & 2.133 \\
\bottomrule
\end{tabular}
}
\caption{Closed-loop results on the same 50 tasks. Mean distance reduction is the initial minus final planar distance to the target box, averaged over all episodes.}
\label{tab:closed_loop}
\end{table}

In Table~\ref{tab:closed_loop}, fine-tuning increases the paired success rate by 20 percentage points (95\% CI of 10--30 points from 10,000 resamples of scenes). With the same fine-tuned model, single-image inference and graph scoring averaged 6.2\,s per task versus 118\,s per Isaac Sim episode---nearly 20$\times$ faster per evaluation unit.

\begin{figure}[!ht]
\vspace{-6pt}
\centering
\includegraphics[width=0.70\linewidth]{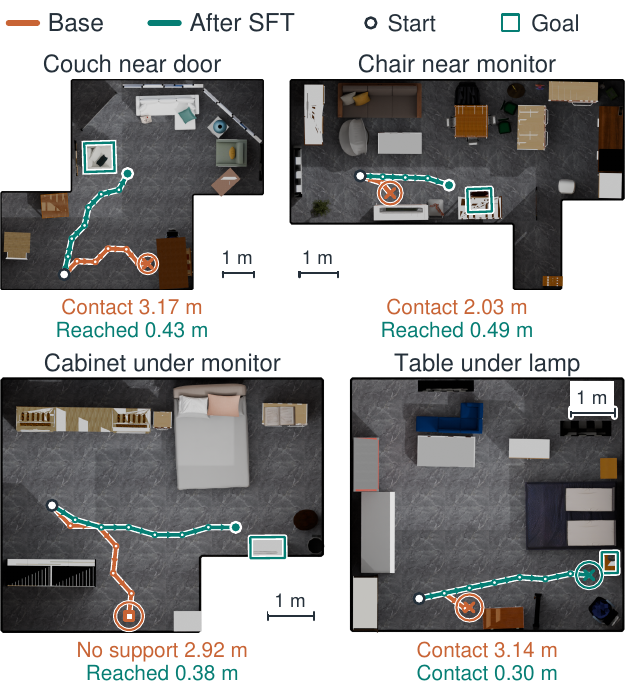}
\caption{Rings: 0.6\,m robot diameter; labels: final goal distance.}
\label{fig:closed_loop_rollout}
\vspace{-6pt}
\end{figure}

Figure~\ref{fig:closed_loop_rollout} compares paired rollouts across four scenes: the base model repeatedly collides with furniture or steps off supported floor (terminating 2.03--3.17\,m from the target), whereas the EgoPath-SFT model navigates around obstacles to reach within the 0.6\,m goal radius in three scenes (0.38--0.49\,m) while still colliding in tight clutter (\emph{Table under lamp}, 0.30\,m).

\FloatBarrier

\section{Conclusion}

EgoPathBench evaluates whether VLMs turn first-person observations and goals into feasible, goal-reaching routes across five scene-grounded tasks.

Across ten VLMs, the highest EgoPath Score is 35.6; the leader reaches 41.7\% Point Path success, but drops to 10.0\% on Embodied and Intent Path. Fine-tuning Qwen3.5-4B on EgoPath-SFT raises its score from 3.9 to 38.9, improves closed-loop execution by 20 points, and transfers to three external benchmarks. Bridging isolated spatial judgments with sustained geometric and stopping consistency in multi-step embodied decisions remains key for future VLMs.

\label{content:end}

\section*{LLM Usage Statement}
OpenAI Codex was used to assist with writing and polishing the manuscript, and GPT-5.5 was used to generate synthetic dataset content. For the latter, the Data Construction and Annotation section and the Training-resource evaluation paragraph in Appendix~A describe the use of GPT-5.5 to verbalize fixed targets, candidates, feasibility labels, legal edges, and reference routes into Spatial CoT supervision. As described there, target identity and geometric annotations were fixed before language generation, and exported answers were checked against the formal annotations. The authors take responsibility for the final content of this work, including AI-assisted text and dataset artifacts.

{
\small
\bibliographystyle{ieeenat_fullname}
\bibliography{references}

\begin{thebibliography}{39}
\providecommand{\natexlab}[1]{#1}
\providecommand{\url}[1]{\texttt{#1}}
\expandafter\ifx\csname urlstyle\endcsname\relax
  \providecommand{\doi}[1]{doi: #1}\else
  \providecommand{\doi}{doi: \begingroup \urlstyle{rm}\Url}\fi

\bibitem[Anderson et~al.(2018)Anderson, Wu, Teney, Bruce, Johnson, Sunderhauf,
  Reid, Gould, and van~den Hengel]{r2r2018}
Peter Anderson, Qi Wu, Damien Teney, Jake Bruce, Mark Johnson, Niko Sunderhauf,
  Ian Reid, Stephen Gould, and Anton van~den Hengel.
\newblock Vision-and-language navigation: Interpreting visually-grounded
  navigation instructions in real environments.
\newblock In \emph{Proceedings of the IEEE Conference on Computer Vision and
  Pattern Recognition}, 2018.

\bibitem[Baruch et~al.(2021)Baruch, Chen, Dehghan, Dimry, Feigin, Fu, Gebauer,
  Joffe, Kurz, Schwartz, and Shulman]{arkitscenes2021}
Gilad Baruch, Zhuoyuan Chen, Afshin Dehghan, Tal Dimry, Yuri Feigin, Peter Fu,
  Thomas Gebauer, Brandon Joffe, Daniel Kurz, Arik Schwartz, and Elad Shulman.
\newblock Arkitscenes: A diverse real-world dataset for 3d indoor scene
  understanding using mobile rgb-d data.
\newblock In \emph{Advances in Neural Information Processing Systems Datasets
  and Benchmarks Track}, 2021.

\bibitem[Cai et~al.(2025)Cai, He, Wang, Guo, Yau, and Lv]{clcotnav2025}
Yuxin Cai, Xiangkun He, Maonan Wang, Hongliang Guo, Wei-Yun Yau, and Chen Lv.
\newblock Cl-cotnav: Closed-loop hierarchical chain-of-thought for zero-shot
  object-goal navigation with vision-language models.
\newblock \emph{arXiv preprint arXiv:2504.09000}, 2025.

\bibitem[Chang et~al.(2017)Chang, Dai, Funkhouser, Halber, Nie{\ss}ner, Savva,
  Song, Zeng, and Zhang]{matterport3d2017}
Angel Chang, Angela Dai, Thomas Funkhouser, Maciej Halber, Matthias
  Nie{\ss}ner, Manolis Savva, Shuran Song, Andy Zeng, and Yinda Zhang.
\newblock Matterport3d: Learning from rgb-d data in indoor environments.
\newblock In \emph{Proceedings of the International Conference on 3D Vision},
  2017.

\bibitem[Chen et~al.(2024)Chen, Xu, Kirmani, Ichter, Driess, Florence, Sadigh,
  Guibas, and Xia]{spatialvlm2024}
Boyuan Chen, Zhuo Xu, Sean Kirmani, Brian Ichter, Danny Driess, Pete Florence,
  Dorsa Sadigh, Leonidas Guibas, and Fei Xia.
\newblock Spatialvlm: Endowing vision-language models with spatial reasoning
  capabilities.
\newblock In \emph{Proceedings of the IEEE/CVF Conference on Computer Vision
  and Pattern Recognition}, 2024.

\bibitem[Chen et~al.(2025)Chen, An, Huang, Xu, Su, Ling, Reid, and
  Wang]{canav2025}
Kehan Chen, Dong An, Yan Huang, Rongtao Xu, Yifei Su, Yonggen Ling, Ian Reid,
  and Liang Wang.
\newblock Constraint-aware zero-shot vision-language navigation in continuous
  environments.
\newblock \emph{IEEE Transactions on Pattern Analysis and Machine
  Intelligence}, 2025.

\bibitem[Cheng et~al.(2024)Cheng, Guo, Wu, Fang, Li, Liu, and
  Liu]{egothink2024}
Sijie Cheng, Zhicheng Guo, Jingwen Wu, Kechen Fang, Peng Li, Huaping Liu, and
  Yang Liu.
\newblock Egothink: Evaluating first-person perspective thinking capability of
  vision-language models.
\newblock \emph{arXiv preprint arXiv:2311.15596}, 2024.

\bibitem[Dai et~al.(2017)Dai, Chang, Savva, Halber, Funkhouser, and
  Nie{\ss}ner]{scannet2017}
Angela Dai, Angel~X. Chang, Manolis Savva, Maciej Halber, Thomas Funkhouser,
  and Matthias Nie{\ss}ner.
\newblock Scannet: Richly-annotated 3d reconstructions of indoor scenes.
\newblock In \emph{Proceedings of the IEEE Conference on Computer Vision and
  Pattern Recognition}, 2017.

\bibitem[Du et~al.(2024)Du, Wu, Li, Huang, and Wei]{embspatial2024}
Mengfei Du, Binhao Wu, Zejun Li, Xuanjing Huang, and Zhongyu Wei.
\newblock Embspatial-bench: Benchmarking spatial understanding for embodied
  tasks with large vision-language models.
\newblock In \emph{Proceedings of the 62nd Annual Meeting of the Association
  for Computational Linguistics}, 2024.

\bibitem[Goetting et~al.(2025)Goetting, Singh, and
  Loquercio]{goetting2025vlmnav}
Dylan Goetting, Himanshu~Gaurav Singh, and Antonio Loquercio.
\newblock End-to-end navigation with vision-language models: Transforming
  spatial reasoning into question-answering.
\newblock In \emph{Proceedings of the International Conference on
  Neuro-symbolic Systems}, pages 22--35, 2025.

\bibitem[Krantz et~al.(2020)Krantz, Wijmans, Majumdar, Batra, and
  Lee]{vlnce2020}
Jacob Krantz, Erik Wijmans, Arjun Majumdar, Dhruv Batra, and Stefan Lee.
\newblock Beyond the nav-graph: Vision-and-language navigation in continuous
  environments.
\newblock In \emph{Proceedings of the European Conference on Computer Vision},
  2020.

\bibitem[Ku et~al.(2020)Ku, Anderson, Patel, Ie, and Baldridge]{rxr2020}
Alexander Ku, Peter Anderson, Roma Patel, Eugene Ie, and Jason Baldridge.
\newblock Room-across-room: Multilingual vision-and-language navigation with
  dense spatiotemporal grounding.
\newblock In \emph{Proceedings of the 2020 Conference on Empirical Methods in
  Natural Language Processing}, 2020.

\bibitem[Li et~al.(2025{\natexlab{a}})Li, Li, Wang, Yan, Zhang, Chen, Hou,
  Jiang, Zhang, Shen, Lu, and Zhuang]{viewspatial2025}
Dingming Li, Hongxing Li, Zixuan Wang, Yuchen Yan, Hang Zhang, Siqi Chen,
  Guiyang Hou, Shengpei Jiang, Wenqi Zhang, Yongliang Shen, Weiming Lu, and
  Yueting Zhuang.
\newblock Viewspatial-bench: Evaluating multi-perspective spatial localization
  in vision-language models.
\newblock \emph{arXiv preprint arXiv:2505.21500}, 2025{\natexlab{a}}.

\bibitem[Li et~al.(2025{\natexlab{b}})Li, Li, Dao, Zhou, Huang, Ma, Qiao, Mai,
  Lee, Chen, Wang, Yang, Wang, Tan, Li, Bansal, Ni, and Kong]{industrynav2025}
Yifan Li, Lichi Li, Anh Dao, Xinyu Zhou, Wenjun Huang, Tianyi Ma, Yicheng Qiao,
  Zheda Mai, Daeun Lee, Zichen Chen, Pan Wang, Lehan Yang, Tianlong Wang, Zhen
  Tan, Sheng Li, Mohit Bansal, Yang Ni, and Yu Kong.
\newblock Industrynav: Exploring spatial reasoning of embodied agents in
  dynamic industrial navigation.
\newblock \emph{arXiv preprint arXiv:2511.17384}, 2025{\natexlab{b}}.

\bibitem[Li et~al.(2026)Li, Li, Shi, Luo, Cai, Yang, and Qin]{agenticnav2026}
Yijian Li, Changze Li, Hantian Shi, Jiaying Luo, Jiyuan Cai, Ming Yang, and
  Tong Qin.
\newblock Agenticnav: Zero-shot vision-and-language navigation as a
  tool-calling harness.
\newblock \emph{arXiv preprint arXiv:2606.10577}, 2026.

\bibitem[Liu et~al.(2025)Liu, Huang, Zhang, and Tang]{navr12025}
Qingxiang Liu, Ting Huang, Zeyu Zhang, and Hao Tang.
\newblock Nav-r1: Reasoning and navigation in embodied scenes.
\newblock \emph{arXiv preprint arXiv:2509.10884}, 2025.

\bibitem[Long et~al.(2024)Long, Cai, Wang, Zhan, and Dong]{instructnav2024}
Yuxing Long, Wenzhe Cai, Hongcheng Wang, Guanqi Zhan, and Hao Dong.
\newblock Instructnav: Zero-shot system for generic instruction navigation in
  unexplored environment.
\newblock In \emph{Proceedings of the Conference on Robot Learning}, 2024.

\bibitem[Ma et~al.(2025)Ma, Chen, Zhang, de~Melo, Yuille, and
  Chen]{threedsrbench2025}
Wufei Ma, Haoyu Chen, Guofeng Zhang, Celso~M. de Melo, Alan Yuille, and Jieneng
  Chen.
\newblock 3dsrbench: A comprehensive 3d spatial reasoning benchmark.
\newblock In \emph{Proceedings of the IEEE/CVF International Conference on
  Computer Vision}, 2025.

\bibitem[Majumdar et~al.(2024)]{openeqa2024}
Arjun Majumdar et~al.
\newblock Openeqa: Embodied question answering in the era of foundation models.
\newblock In \emph{Proceedings of the IEEE/CVF Conference on Computer Vision
  and Pattern Recognition}, 2024.

\bibitem[Nasiriany et~al.(2024)Nasiriany, Xia, Yu, Xiao, Liang, Dasgupta, Xie,
  Driess, Wahid, Xu, Vuong, Zhang, Lee, Lee, Xu, Kirmani, Zhu, Zeng, Hausman,
  Heess, Finn, Levine, and Ichter]{nasiriany2024pivot}
Soroush Nasiriany, Fei Xia, Wenhao Yu, Ted Xiao, Jacky Liang, Ishita Dasgupta,
  Annie Xie, Danny Driess, Ayzaan Wahid, Zhuo Xu, Quan Vuong, Tingnan Zhang,
  Tsang-Wei~Edward Lee, Kuang-Huei Lee, Peng Xu, Sean Kirmani, Yuke Zhu, Andy
  Zeng, Karol Hausman, Nicolas Heess, Chelsea Finn, Sergey Levine, and Brian
  Ichter.
\newblock {PIVOT}: Iterative visual prompting elicits actionable knowledge for
  vlms.
\newblock In \emph{Proceedings of the 41st International Conference on Machine
  Learning}, pages 37321--37341, 2024.

\bibitem[Qi et~al.(2020)Qi, Wu, Anderson, Wang, Wang, Shen, and van~den
  Hengel]{reverie2020}
Yuankai Qi, Qi Wu, Peter Anderson, Xin Wang, William~Yang Wang, Chunhua Shen,
  and Anton van~den Hengel.
\newblock Reverie: Remote embodied visual referring expression in real indoor
  environments.
\newblock In \emph{Proceedings of the IEEE/CVF Conference on Computer Vision
  and Pattern Recognition}, 2020.

\bibitem[Qiao et~al.(2025{\natexlab{a}})Qiao, Hong, Lyu, An, Zhang, Xie, Wang,
  and Wu]{navbench2025}
Yanyuan Qiao, Haodong Hong, Wenqi Lyu, Dong An, Siqi Zhang, Yutong Xie, Xinyu
  Wang, and Qi Wu.
\newblock Navbench: Probing multimodal large language models for embodied
  navigation.
\newblock In \emph{Advances in Neural Information Processing Systems},
  2025{\natexlab{a}}.

\bibitem[Qiao et~al.(2025{\natexlab{b}})Qiao, Lyu, Wang, Wang, Li, Zhang, Tan,
  and Wu]{opennav2025}
Yanyuan Qiao, Wenqi Lyu, Hui Wang, Zixu Wang, Zerui Li, Yuan Zhang, Mingkui
  Tan, and Qi Wu.
\newblock Open-nav: Exploring zero-shot vision-and-language navigation in
  continuous environment with open-source llms.
\newblock In \emph{Proceedings of the IEEE International Conference on Robotics
  and Automation (ICRA)}, 2025{\natexlab{b}}.

\bibitem[Sheng et~al.(2026)Sheng, Wang, Dai, Li, Qin, He, Liu, and
  Chen]{p2dnav2026}
Kai Sheng, Liuyi Wang, Haojie Dai, Jinlong Li, Yongrui Qin, Zongtao He, Chengju
  Liu, and Qijun Chen.
\newblock {P2DNav}: Panorama-to-downview reasoning for zero-shot
  vision-and-language navigation.
\newblock \emph{arXiv preprint arXiv:2605.19634}, 2026.

\bibitem[Shi et~al.(2025)Shi, Li, Lyu, Xia, Dayoub, Qiao, and Wu]{smartway2025}
Xiangyu Shi, Zerui Li, Wenqi Lyu, Jiatong Xia, Feras Dayoub, Yanyuan Qiao, and
  Qi Wu.
\newblock Smartway: Enhanced waypoint prediction and backtracking for zero-shot
  vision-and-language navigation.
\newblock In \emph{Proceedings of the IEEE/RSJ International Conference on
  Intelligent Robots and Systems}, 2025.

\bibitem[Song et~al.(2025)Song, Chen, Liu, Chen, Li, and Lin]{lhprvln2025}
Xinshuai Song, Weixing Chen, Yang Liu, Weikai Chen, Guanbin Li, and Liang Lin.
\newblock Towards long-horizon vision-language navigation: Platform, benchmark
  and method.
\newblock In \emph{Proceedings of the IEEE/CVF Conference on Computer Vision
  and Pattern Recognition}, 2025.

\bibitem[Su et~al.(2026)Su, Chen, Liu, Ma, Di, Krishna, and
  Froehlich]{capnav2026}
Xia Su, Ruiqi Chen, Benlin Liu, Jingwei Ma, Zonglin Di, Ranjay Krishna, and Jon
  Froehlich.
\newblock Capnav: Benchmarking vision language models on capability-conditioned
  indoor navigation.
\newblock \emph{arXiv preprint arXiv:2602.18424}, 2026.

\bibitem[Wald et~al.(2019)Wald, Dhamo, Navab, and Tombari]{threerscan2019}
Johanna Wald, Helisa Dhamo, Nassir Navab, and Federico Tombari.
\newblock Rio: 3d object instance re-localization in changing indoor
  environments.
\newblock In \emph{Proceedings of the IEEE/CVF International Conference on
  Computer Vision}, 2019.

\bibitem[Wang et~al.(2024)Wang, Ming, Shi, Vineet, Wang, Li, and
  Joshi]{spatialeval2024}
Jiayu Wang, Yifei Ming, Zhenmei Shi, Vibhav Vineet, Xin Wang, Yixuan Li, and
  Neel Joshi.
\newblock Is a picture worth a thousand words? delving into spatial reasoning
  for vision language models.
\newblock In \emph{Advances in Neural Information Processing Systems
  (NeurIPS)}, 2024.

\bibitem[Wang et~al.(2025)Wang, Fang, Wang, Feng, Tan, Zhang, Liu, Ji, and
  Xu]{dreamnav2025}
Yunheng Wang, Yuetong Fang, Taowen Wang, Yixiao Feng, Yawen Tan, Shuning Zhang,
  Peiran Liu, Yiding Ji, and Renjing Xu.
\newblock Dreamnav: A trajectory-based imaginative framework for zero-shot
  vision-and-language navigation.
\newblock \emph{arXiv preprint arXiv:2509.11197}, 2025.

\bibitem[Windecker et~al.(2025)Windecker, Patel, Reuss, Schwarzkopf, Cadena,
  Lioutikov, Hutter, and Frey]{navitrace2025}
Tim Windecker, Manthan Patel, Moritz Reuss, Richard Schwarzkopf, Cesar Cadena,
  Rudolf Lioutikov, Marco Hutter, and Jonas Frey.
\newblock Navitrace: Evaluating embodied navigation of vision-language models.
\newblock \emph{arXiv preprint arXiv:2510.26909}, 2025.

\bibitem[Yang et~al.(2023)Yang, Zhang, Li, Zou, Li, and Gao]{yang2023setofmark}
Jianwei Yang, Hao Zhang, Feng Li, Xueyan Zou, Chunyuan Li, and Jianfeng Gao.
\newblock Set-of-mark prompting unleashes extraordinary visual grounding in
  gpt-4v.
\newblock \emph{arXiv preprint arXiv:2310.11441}, 2023.

\bibitem[Yang et~al.(2025)Yang, Yang, Gupta, Han, Fei-Fei, and
  Xie]{vsibench2025}
Jihan Yang, Shusheng Yang, Anjali~W. Gupta, Rilyn Han, Li Fei-Fei, and Saining
  Xie.
\newblock Thinking in space: How multimodal large language models see,
  remember, and recall spaces.
\newblock In \emph{Proceedings of the IEEE/CVF Conference on Computer Vision
  and Pattern Recognition}, 2025.

\bibitem[Yokoyama et~al.(2024)Yokoyama, Ha, Batra, Wang, and Bucher]{vlfm2024}
Naoki Yokoyama, Sehoon Ha, Dhruv Batra, Jiuguang Wang, and Bernadette Bucher.
\newblock {VLFM}: Vision-language frontier maps for zero-shot semantic
  navigation.
\newblock In \emph{Proceedings of the IEEE International Conference on Robotics
  and Automation}, 2024.

\bibitem[Zhang et~al.(2024)Zhang, Wang, Xu, Zhou, Hong, Fang, Wu, Zhang, and
  Wang]{navid2024}
Jiazhao Zhang, Kunyu Wang, Rongtao Xu, Gengze Zhou, Yicong Hong, Xiaomeng Fang,
  Qi Wu, Zhizheng Zhang, and He Wang.
\newblock Navid: Video-based vlm plans the next step for vision-and-language
  navigation.
\newblock \emph{Robotics: Science and Systems}, 2024.

\bibitem[Zhang et~al.(2026)Zhang, Zhang, Peng, Liu, Wang, Long, Huang, Li,
  Duan, Shen, and Dong]{embodied3dbench2026}
Jiyao Zhang, Mingxu Zhang, Yitong Peng, Haoxuan Liu, Chenshuo Wang, Yuxing
  Long, Haoyang Huang, Dongjiang Li, Nan Duan, Hui Shen, and Hao Dong.
\newblock Embodied3dbench: Benchmarking low-level embodied spatial intelligence
  of vision language models.
\newblock \emph{arXiv preprint arXiv:2605.29074}, 2026.

\bibitem[Zheng et~al.(2024)Zheng, Huang, Zhao, Zhong, and Wang]{navillm2024}
Duo Zheng, Shijia Huang, Lin Zhao, Yiwu Zhong, and Liwei Wang.
\newblock Towards learning a generalist model for embodied navigation.
\newblock In \emph{Proceedings of the IEEE/CVF Conference on Computer Vision
  and Pattern Recognition}, 2024.

\bibitem[Zhong et~al.(2025)Zhong, Cao, Jin, Li, Cai, Lin, Wang, Lyu, Wang, Dai,
  Xu, and Pang]{internscenes2025}
Weipeng Zhong, Peizhou Cao, Yichen Jin, Luo Li, Wenzhe Cai, Jingli Lin, Hanqing
  Wang, Zhaoyang Lyu, Tai Wang, Bo Dai, Xudong Xu, and Jiangmiao Pang.
\newblock Internscenes: A large-scale simulatable indoor scene dataset with
  realistic layouts.
\newblock In \emph{Advances in Neural Information Processing Systems Datasets
  and Benchmarks Track}, 2025.

\bibitem[Zhou et~al.(2024)Zhou, Hong, and Wu]{navgpt2024}
Gengze Zhou, Yicong Hong, and Qi Wu.
\newblock Navgpt: Explicit reasoning in vision-and-language navigation with
  large language models.
\newblock In \emph{Proceedings of the AAAI Conference on Artificial
  Intelligence}, 2024.

\end{thebibliography}
}

\clearpage
\appendix
\setcounter{figure}{0}
\renewcommand{\thefigure}{S\arabic{figure}}
\setcounter{table}{0}
\renewcommand{\thetable}{S\arabic{table}}
\renewcommand{\theHfigure}{appendix.\arabic{figure}}
\renewcommand{\theHtable}{appendix.\arabic{table}}
\section{Evaluation Protocol and Reproducibility}
\label{app:evaluation_protocol}

\paragraph{Reproducibility.}
Detailed implementation information is provided in the appendices. The code and data will be made publicly available upon acceptance of the paper.

\begin{table*}[t]
\centering
{
\footnotesize
\setlength{\tabcolsep}{1.7pt}
\begin{tabular}{@{}p{0.21\linewidth}lllp{0.29\linewidth}r@{}}
\toprule
Task & Goal & Evaluation graph & Output & Scored constraints & N \\
\midrule
Point Trav. & -- & Point & Set & Traversability & 146 \\
Embodied Trav. & -- & Embodied & Set & Feasibility & 146 \\
\midrule
Point Path & Explicit & Point & Route & Edges, endpoint & 309 \\
Embodied Path & Explicit & Embodied & Route & Footprint, edges, goal & 309 \\
Intent Path & Intent & Embodied & Route & Intent, footprint, edges & 201 \\
\bottomrule
\end{tabular}
}
\caption{Five EgoPathBench tasks. The shared first-person waypoint interface is specialized by goal specification, agent geometry, output structure, evaluated constraints, and benchmark question count.}
\label{tab:task_taxonomy}
\end{table*}

Table~\ref{tab:supp_task_interfaces} summarizes the observation, action, and scoring contract for each task. Traversability tasks require a set of locally feasible waypoint actions. Route tasks require an ordered sequence beginning at the specified start; every consecutive edge must be legal for the task's agent geometry, and the final waypoint must belong to the acceptable goal set.

\begin{table*}[!ht]
\centering
\small
\setlength{\tabcolsep}{4pt}
\begin{tabular}{@{}>{\raggedright\arraybackslash}p{0.14\textwidth}>{\raggedright\arraybackslash}p{0.19\textwidth}>{\raggedright\arraybackslash}p{0.14\textwidth}>{\raggedright\arraybackslash}p{0.18\textwidth}>{\raggedright\arraybackslash}p{0.27\textwidth}@{}}
\toprule
Task & Decision input & Output & Task specification & Primary scoring condition \\
\midrule
Point Traversability & RGB with shared visible candidates & ID set & Walkable candidates & Candidate-wise traversability classification \\
Embodied Traversability & RGB with shared visible candidates & ID set & 0.6\,m body; no goal & Candidate-wise body-feasible classification \\
Point Path & RGB, start ID, explicit target & Ordered route & Explicit goal & Valid IDs, required start, legal edges, acceptable endpoint \\
Embodied Path & RGB, start ID, explicit target & Ordered route & 0.6 m body; explicit goal & Point-Path checks plus footprint-aware edge legality \\
Intent Path & RGB, start ID, natural-language intent & Ordered route & 0.6 m body; resolved goal & Intent-consistent endpoint and embodied route legality \\
\bottomrule
\end{tabular}
\caption{Interface and scoring contract for the five EgoPathBench tasks. All outputs use visible display IDs from the current first-person image.}
\label{tab:supp_task_interfaces}
\end{table*}

\paragraph{Prompt and output protocol.}
We use everyday navigation requests to test whether models spontaneously infer an appropriate stopping location from the scene and goal, without an explicit stopping reminder. Every question stores the system prompt, user prompt, image, and visible-waypoint record used for evaluation. Table~\ref{tab:supp_prompt_contract} gives the released prompt templates. Bracketed fields are filled from the fixed question record; the requested answer is always one JSON array of visible display IDs.

\begin{table*}[!ht]
\centering
\small
\setlength{\tabcolsep}{4pt}
\begin{tabular}{@{}>{\raggedright\arraybackslash}p{0.12\textwidth}p{0.27\textwidth}p{0.55\textwidth}@{}}
\toprule
Task & System prompt & User-prompt template \\
\midrule
Point Traversability & You are a navigation perception assistant. & The image shows numbered candidate points. Output a JSON array of display IDs that are walkable. \\
Embodied Traversability & You are a navigation perception assistant. & The image shows numbered candidate points. The robot has diameter 0.6m. Output a JSON array of display IDs that are walkable for the robot. \\
Point Path & You are a navigation planning assistant. & The image shows numbered candidate points. Start at display ID 1. Navigate to the [explicit target]. Output a JSON array of display IDs representing a valid path. \\
Embodied Path & You are an embodied navigation robot. Plan a collision-free path for your body using the numbered candidate points shown in the image. & The image shows numbered candidate points. You are the robot, and your body diameter is 0.6 m. Start at display ID 1. Navigate to the [explicit target]. Return a JSON array of display IDs representing a valid collision-free path for the robot. \\
Intent Path & You are an embodied navigation robot. A user gives you a short request that implies the kind of object they want. Infer the target family from the request, resolve the final grounded target in the scene, and plan a collision-free path for your body. & The image shows numbered candidate points. You are the robot, and your body diameter is 0.6 m. Start at display ID 1. A person says: ``[intent request].'' Return a JSON array of display IDs representing a valid collision-free path to the resolved target. \\
\bottomrule
\end{tabular}
\caption{Released prompt templates for the five tasks. Bracketed fields are populated from the fixed question record.}
\label{tab:supp_prompt_contract}
\end{table*}

\paragraph{Route evaluator.}
The released sidecar applies ordered candidate-validity, start, direct-edge, and goal-membership checks; SPL is computed only after a route satisfies this success contract. Table~\ref{tab:evaluation_contract_details} lists the complete check sequence.

\begin{center}
\small
\setlength{\tabcolsep}{2pt}
\begin{tabular}{@{}p{0.19\linewidth}p{0.52\linewidth}p{0.20\linewidth}@{}}
\toprule
Stage & Pass condition & Outcome \\
\midrule
Parse & One ordered ID list is recovered. & Format \\
Candidate & Every ID is visible and allowed. & Invalid ID \\
Start & The first ID matches the required start. & Wrong start \\
Edge & Every consecutive pair is a legal direct edge. & Illegal edge \\
Endpoint & The final ID is an acceptable goal. & Wrong goal \\
Efficiency & A successful route is compared with shortest references. & SPL \\
\bottomrule
\end{tabular}
\captionof{table}{Ordered evaluation contract for route-bearing tasks. SPL is applied after route success.}
\label{tab:evaluation_contract_details}
\end{center}

\paragraph{Navigation maps.}
The Point and Embodied feasibility graphs (also referred to as navigation maps) use a height-filtered 2D footprint approximation. Obstacle geometry intersecting the 0.10--1.40 m height interval (modeling a service robot's vertical clearance envelope while filtering floor noise and ceiling fixtures) is projected onto a grid with 0.05 m cells. The Embodied map uses a 0.30 m radius with no additional margin to expand obstacles and shrink the walkable floor region. When an obstacle mesh is unavailable, its expanded oriented bounding rectangle is used. The Point map uses the same geometric representation without footprint expansion. An edge between two visible waypoints is legal if the straight-line corridor swept by the agent's footprint (0.30 m radius for Embodied and Intent; point segment for Point) is collision-free.

\paragraph{Acceptable endpoints.}
For Embodied Path and Intent Path, we compute the 2D distance from each embodied-feasible candidate waypoint to the target's oriented bounding rectangle. Let $d_{\min}$ be the smallest such distance. The acceptable endpoint set contains all embodied-feasible candidates within $d_{\min}+0.10$ m of the rectangle. For Point Path, the initial threshold is computed from point-feasible candidates in the same way and is increased, when needed, to include the nearest embodied-feasible candidate. Because a finite-sized robot requires clearance that a point agent does not, we define endpoints over each agent's feasible candidates and extend the Point region to retain the nearest robot-feasible stopping candidate in both tasks. This extension applies to 253 of the 309 Point Path questions. A predicted route reaches the goal when its final waypoint belongs to the corresponding acceptable endpoint set.

\paragraph{Metric definitions.}
Task-feasible candidates are the positive class for traversability. Let $P$ and $R$ denote precision and recall, and let $V_i$ indicate that route prediction $i$ is parseable, uses valid IDs, begins at the required start, and contains only legal consecutive edges. Let $S_i$ additionally require an acceptable endpoint. For $N$ questions,
\begin{align*}
\mathrm{BA} &= \frac{\mathrm{TPR}+\mathrm{TNR}}{2},
& F1 &= \frac{2PR}{P+R}, \\
\mathrm{VPR} &= \frac{1}{N}\sum_i V_i,
& \mathrm{SR} &= \frac{1}{N}\sum_i S_i, \\
\mathrm{SPL} &= \frac{1}{N}\sum_i S_i
\frac{\ell_i}{\max(\ell_i,p_i)},
\end{align*}
where $\ell_i$ is the shortest legal reference length to an acceptable goal and $p_i$ is the predicted route length. Thus, unsuccessful routes receive zero SPL. With component metrics expressed in $[0,1]$, the reported aggregate is
\begin{align*}
\mathrm{EgoPath\ Score}=\frac{100}{5}\big[ &(2\mathrm{BA}_{\mathrm{PT}}-1)
+(2\mathrm{BA}_{\mathrm{ET}}-1) \\
&+\mathrm{SR}_{\mathrm{PP}}+\mathrm{SR}_{\mathrm{EP}}
+\mathrm{SR}_{\mathrm{IP}}\big].
\end{align*}

\paragraph{Stopping-instruction control.}
We compare the original prompts with a condition that appends: \emph{Stop at the numbered candidate point that is feasible for the specified agent and closest to the target object.} Both conditions use the same 50 fixed questions (20 Point, 20 Embodied, 10 Intent), identical images, and the benchmark's evaluation rules. Table~\ref{tab:stopping_instruction} in the main text reports endpoint hits and route success. The stopping reminder increases endpoint hits from 13/50 to 20/50 and successful routes from 7/50 to 12/50.

\paragraph{Foundation VLM evaluation.}
The ten endpoint identifiers are \emph{gpt-6-astra}, \emph{claude-opus-4-8}, \emph{gemini-3.1-pro-preview}, \emph{gpt-5.5}, \emph{grok-4.3-fast}, \emph{kimi-k2.6}, \emph{meta/llama-4-maverick-17b-128e-instruct}, \emph{MiniMax-M3}, \emph{mistralai/mistral-large-3-675b-instruct-2512}, and \emph{qwen3.6-plus}. We request an 8,192-token completion limit for every model. We use temperature 0 and top-p 1 where supported, with provider defaults for controls that are unavailable. Each leaderboard entry covers the same 1,111 questions with the same task prompts and JSON output protocol.

\paragraph{Training-resource evaluation.}
We train Qwen3.5-4B with LoRA rank 8, alpha 16, and zero dropout while freezing the vision tower. Stage one uses per-device batch size 1, gradient accumulation 8, a cosine schedule from $10^{-4}$ with 10\% warmup, two epochs, and seed 42. Continuation from the final adapter uses two further epochs at $5\!\times\!10^{-5}$ with 5\% warmup. Both stages use bfloat16, a 4,096-token cutoff, and maximum image area 262,144 pixels. We evaluate the checkpoint saved at a fixed 3,000-step continued-training budget. EgoPathBench and external evaluation use completion budgets of 8,192 and 4,096 tokens, respectively.

External evaluation uses the VSI-Bench Route Planning subset in both its Full and Debiased settings, together with SpatialEval-VTQA and 3DSRBench. The base model and selected checkpoint are evaluated on the same items and with the same protocol within each setting. VSI-Bench provides the navigation-focused video evaluation, while the other two benchmarks test transfer to complementary spatial tasks.

The EgoPath-SFT export contains one image-grounded conversation for every training question. Its assistant response contains the accepted GPT-5.5-written Spatial CoT followed by the geometry-verified JSON answer; format repair may normalize the surrounding tags and final answer line but does not replace the rationale. Table~\ref{tab:spatial_cot_contract} reports the release audit.

\begin{center}
\small
\begin{tabular}{p{0.55\linewidth}p{0.34\linewidth}}
\toprule
Field & Value \\
\midrule
Complete training rows & 31,852 / 31,852 \\
Missing image / rationale & 0 / 0 \\
Rationale source & GPT-5.5-written, non-template \\
Format and answer audit & Pass \\
\bottomrule
\end{tabular}
\captionof{table}{Audit of the released EgoPath-SFT export.}
\label{tab:spatial_cot_contract}
\end{center}

\paragraph{Closed-loop simulation protocol.}
Closed-loop evaluation runs 50 fixed start--goal tasks across 25 held-out scenes using a 0.6\,m-diameter embodied footprint in Isaac Sim. At each decision step, the simulator projects visible ground candidate waypoints onto the current egocentric RGB view using the same numbered-ID interface as the single-view benchmark, and the controller executes up to 0.6\,m along the first selected segment. Episodes terminate upon collision, loss of floor support, reaching within 0.6\,m of the target's ground-projected bounding box, or exceeding 12 decision steps.

\paragraph{Software, compute, and release.}
Scene rendering and geometric annotation use Blender 4.4 and Python 3.12. Model training uses Python 3.10.20, PyTorch 2.12.0 with CUDA 13.0, Transformers 5.2.0, PEFT 0.15.1, Accelerate 1.6.0, and LLaMA-Factory on two NVIDIA A100 80GB GPUs. Upon publication, we will release the generated questions, annotations, training supervision, task prompts, construction and evaluation code, model configurations, and fixed prediction records under licenses permitting research use and consistent with the terms of the upstream scene assets.

\paragraph{Aggregate-score sensitivity.}
Replacing route SR with SPL gives Spearman $\rho=1.00$ against the primary ten-model ordering. Weighting traversability and route tasks equally as two families gives $\rho=1.00$.

\section{Geometric Evaluation Validation}
\label{app:geometric_validation}

\paragraph{Human-audit sampling.}
For the human audit in Section~\ref{sec:geometric_answerability}, we select 40 subpaths from model predictions on the Embodied Path task, comprising 20 geometrically valid and 20 invalid subpaths. The sample contains 14 subpaths with two waypoints, 14 with three waypoints, and 12 with four waypoints. All four participants judge the same 40 subpaths using the RGB-only protocol described in that section. All participants in both human studies provided oral informed consent; institutional ethics review was not sought because the authors assessed these rendered-scene judgment tasks as low risk.

\subsection{Reference-Route Admission and Target Visibility}
\label{app:route_visibility}

\paragraph{Reference-route admission.}
Reference routes are checked for projection into the current view and against the visible-obstacle mask before admission. Table~\ref{tab:single_view_support} reports these checks for all 819 route questions: every dense reference point projects into the view, and both sparse and dense routes pass the mask check. Target bounding-box size and visible-surface statistics characterize the corresponding target observations.

\begin{center}
\small
\setlength{\tabcolsep}{3pt}
\begin{tabular}{@{}lrrrrr@{}}
\toprule
Task & $N$ & Path proj. & Mask pass & BBox px & Surface \\
\midrule
Point & 309 & 100.0 & 100.0 & 37/82 & 32.7/56.2 \\
Embodied & 309 & 100.0 & 100.0 & 37/82 & 32.7/56.2 \\
Intent & 201 & 100.0 & 100.0 & 49/93 & 31.2/53.1 \\
\bottomrule
\end{tabular}
\captionof{table}{Reference-route admission checks for 819 route questions. Path projection covers every dense reference point; mask pass checks sparse and dense routes against the visible-obstacle mask. Target columns report P10/median.}
\label{tab:single_view_support}
\end{center}

\paragraph{Target-visibility analysis.}
Table~\ref{tab:visibility_sensitivity} compares the full set with the 118 Point, 118 Embodied, and 83 Intent questions where the target bounding-box short side is at least 64 pixels and at least 50\% of its surface is visible. On these subsets, endpoint hit exceeds full-route success by 4.6--13.5 percentage points. Illegal-edge rates are 53.6\%, 89.7\%, and 88.6\%.

\begin{center}
\small
\setlength{\tabcolsep}{4pt}
\begin{tabular}{@{}llrrrr@{}}
\toprule
Task & View & $N$ & End/Full & Gap & Illegal \\
\midrule
Point & All & 309 & 31.4/19.2 & 12.2 & 47.3 \\
Point & Strict & 118 & 30.2/16.7 & 13.5 & 53.6 \\
\midrule
Embodied & All & 309 & 5.9/1.9 & 3.9 & 88.7 \\
Embodied & Strict & 118 & 7.0/2.5 & 4.6 & 89.7 \\
\midrule
Intent & All & 201 & 6.3/2.3 & 4.0 & 87.3 \\
Intent & Strict & 83 & 7.3/2.5 & 4.8 & 88.6 \\
\bottomrule
\end{tabular}
\captionof{table}{Visibility sensitivity over ten VLMs (\%). Strict additionally requires target bbox short side $\geq64$ px and visible surface $\geq50\%$. End/Full reports endpoint hit/full-route success.}
\label{tab:visibility_sensitivity}
\end{center}

\subsection{Scene-Consequence Audit and Evaluator Sensitivity}
\label{app:scene_sensitivity}

\paragraph{Scene-consequence audit.}
Across ten VLMs, we audit the illegal edges in 5,957 route predictions containing at least one geometry-defined illegal edge: 1,462 Point, 2,741 Embodied, and 1,754 Intent predictions. We sample each edge at 0.05 m intervals along its centerline for Point paths and across its swept corridor using the formal 0.30 m body radius for Embodied and Intent paths, then search aligned depth and object-index renders for obstruction evidence. At least one explicit registered obstruction is recovered for 5,052 predictions (84.8\%): 4,650 (78.1\%) are supported by depth and a further 402 (6.7\%) by the object-index render alone.

\paragraph{Evaluator sensitivity.}
Table~\ref{tab:geometry_sensitivity} varies the nominal 0.30 m radius, 0.05 m occupancy grid, and 0.10 m goal-distance tolerance. Across variants, edge-label changes range from 0.0\% to 8.5\%, and success-decision changes from 0.2\% to 1.3\%. The ten-model ranking is unchanged (Spearman $\rho=1.00$).

\begin{center}
\small
\setlength{\tabcolsep}{2pt}
\begin{tabular}{@{}lrrrr@{}}
\toprule
Variant & Ref. & Edge flip & Success flip & Rank $\rho$ \\
\midrule
Radius 0.25 m & 100.0 & 2.4 & 0.3 & 1.00 \\
Radius 0.35 m & 67.6 & 8.5 & 0.9 & 1.00 \\
Grid 0.04 m & 91.6 & 5.3 & 1.2 & 1.00 \\
Grid 0.06 m & 89.5 & 6.0 & 1.3 & 1.00 \\
Goal tolerance 0.05 m & 99.8 & 0.0 & 0.2 & 1.00 \\
Goal tolerance 0.15 m & 100.0 & 0.0 & 0.3 & 1.00 \\
\bottomrule
\end{tabular}
\captionof{table}{Geometry sensitivity on 819 route questions and outputs from ten VLMs. Ref. is the fraction retaining a graph solution; edge and success flips are measured against the nominal contract.}
\label{tab:geometry_sensitivity}
\end{center}

\subsection{Endpoint-Rule Controls}
\label{app:endpoint_controls}

\paragraph{Absolute endpoint radius.}
We rescore the frozen outputs of all ten VLMs while holding images, visible candidates, task-specific feasibility graphs, and route-legality checks fixed. An endpoint is accepted when it is feasible for the specified agent and its 2D distance to the target's oriented bounding rectangle is at most 0.6\,m. Point endpoints are selected directly from Point-feasible candidates, without extension to include an Embodied endpoint. Every task retains its full denominator: 309 Point, 309 Embodied, and 201 Intent questions. The two traversability tasks retain their original metrics within the full 1,111-question benchmark.

\begin{center}
\begin{minipage}{\columnwidth}
\centering
\small
\setlength{\tabcolsep}{3pt}
\begin{tabular}{@{}lrrr@{}}
\toprule
Model / endpoint rule & Point & Embodied & Intent \\
\midrule
GPT-6 Astra & 49.84 & 16.18 & 17.41 \\
Gemini 3.1 Pro & 42.39 & 4.21 & 4.98 \\
GPT-5.5 & 37.54 & 1.29 & 1.99 \\
Claude Opus 4.8 & 27.51 & 1.94 & 2.99 \\
MiniMax M3 & 18.12 & 2.27 & 3.48 \\
Qwen 3.6 & 19.74 & 1.29 & 1.49 \\
Mistral Large 3 & 14.89 & 0.00 & 0.50 \\
Llama 4 Maverick & 11.97 & 0.00 & 0.50 \\
Kimi K2.6 & 16.18 & 0.65 & 0.50 \\
Grok 4.3 Fast & 3.24 & 0.00 & 0.00 \\
\midrule
\multicolumn{4}{c}{Pooled SR across ten models} \\
\midrule
Original & 19.22 & 1.94 & 2.29 \\
Point unextended & 15.18 & 1.94 & 2.29 \\
Absolute 0.6 m & 24.14 & 2.78 & 3.38 \\
\bottomrule
\end{tabular}
\captionof{table}{Endpoint-rule controls (SR, \%). Model rows use the absolute 0.6\,m rule with Point extension removed. The pooled rows compare the original protocol, removal of Point extension alone under the original relative-distance rule, and the absolute rule. Each model is scored on all 309 Point, 309 Embodied, and 201 Intent questions.}
\label{tab:absolute_goal_0p6}
\end{minipage}
\end{center}

\paragraph{Results and paired control.}
Table~\ref{tab:absolute_goal_0p6} shows higher Point than Embodied SR for every model under the absolute endpoint rule. Pooled SR is 24.14\% for Point and 2.78\% for Embodied, a 21.36-point gap. Removing only the Point extension from the original relative-distance rule also preserves the gap (15.18\% versus 1.94\%). On the 279 paired questions where both task-specific graphs have a reachable endpoint under the absolute rule, Point and Embodied SR are 25.16\% and 3.08\%, respectively; all ten models retain the same direction of difference. This paired subset is determined by scene geometry and shared across models. These results show that the difficulty of constructing routes under body constraints persists with a shared absolute distance threshold and without Point endpoint extension.

\subsection{Visual-Input Dependence}
\label{app:visual_input}

The input-dependence control keeps the task and candidate count fixed on paired questions and compares the full input with text-only input and a mismatched image--marker overlay. Table~\ref{tab:anti_shortcut} reports route success rates for each condition.

\begin{center}
\small
\setlength{\tabcolsep}{2pt}
\begin{tabular}{@{}lrrr@{}}
\toprule
Model & Full & Text only & Mismatched overlay \\
\midrule
GPT-5.5 & 30.0/0.0/4.0 & 0.0/0.0/0.0 & 10.0/0.0/0.0 \\
Claude Opus 4.8 & 28.0/0.0/0.0 & 0.0/0.0/0.0 & 4.0/0.0/0.0 \\
Qwen 3.6 & 16.0/0.0/0.0 & 0.0/0.0/0.0 & 2.0/0.0/0.0 \\
\bottomrule
\end{tabular}
\captionof{table}{Input-dependence route SR (\%) on fixed paired questions (50 per route task). Cells report Point/Embodied/Intent Path.}
\label{tab:anti_shortcut}
\end{center}

For Point Path, full-input success rates of 30.0\%, 28.0\%, and 16.0\% fall to zero with text-only input and to 10.0\%, 4.0\%, and 2.0\% with mismatched overlays for GPT-5.5, Claude Opus 4.8, and Qwen 3.6, respectively. GPT-5.5's Intent Path success rate falls from 4.0\% to zero under both controls; the remaining Embodied and Intent entries are zero across conditions. These comparisons show that the successful route predictions depend on the paired visual input.

\section{Complete Dataset and Route Statistics}
\label{app:dataset_stats}

\paragraph{Release scale and route structure.}
Table~\ref{tab:dataset_stats} separates release scale from benchmark route structure. The scale block distinguishes scenes, scene--camera views, route units shared by paired questions, target instances, and questions. The structure block shows that route questions are not dominated by direct, unambiguous cases: they contain a median of four same-type objects, Embodied and Intent references require a median of two edges, and embodied routes spend a median 40\% of their length in narrow passages.

\begin{center}
\begin{minipage}{\columnwidth}
\centering
\small
\setlength{\tabcolsep}{3pt}
\textbf{(a) Release scale}\\[-2pt]
\begin{tabular}{@{}lrrrrr@{}}
\toprule
Split & Scenes & Views & Routes & Targets & Qs. \\
\midrule
Train & 2,942 & 6,483 & 7,843 & 6,144 & 31,852 \\
Val & 32 & 200 & 368 & 179 & 1,345 \\
Benchmark & 255 & 365 & 309 & 281 & 1,111 \\
\bottomrule
\end{tabular}
\\[3pt]\textbf{(b) Benchmark route structure}\\[-2pt]
\begin{tabular}{@{}lrrrr@{}}
\toprule
Signal & N & Median & P90 & Max \\
\midrule
Same-type target ambiguity & 819 & 4 & 8 & 16 \\
Reference segments (Point Path) & 309 & 1 & 2 & 3 \\
Reference segments (Embodied Path) & 309 & 2 & 3 & 9 \\
Reference segments (Intent Path) & 201 & 2 & 3 & 9 \\
Reference path length & 819 & 3.2 m & 4.8 m & 8.5 m \\
Embodied narrow-passage fraction & 510 & 40\% & 81\% & 100\% \\
\bottomrule
\end{tabular}
\captionof{table}{Release scale and benchmark route structure. Views are scene--camera pairs; routes are scene--view--route tuples shared by paired route questions.}
\label{tab:dataset_stats}
\end{minipage}
\end{center}

\paragraph{Failure after correct endpoints.}
Table~\ref{tab:route_construct_bridge} selects predictions with a legal first edge and acceptable endpoint. Of these, 38.8\%, 64.1\%, and 60.7\% of Point, Embodied, and Intent routes fail on a later edge. The fewer-waypoint subset shows the same pattern.

\begin{center}
\small
\setlength{\tabcolsep}{3pt}
\textbf{(a) All benchmark questions}\\[-2pt]
\begin{tabular}{@{}lrrr@{}}
\toprule
Task & \shortstack{First edge legal +\\endpoint correct} & \shortstack{Illegal\\later edge} & \shortstack{Complete\\route legal} \\
\midrule
Point & 970 & 376 (38.8\%) & 594 (61.2\%) \\
Embodied & 167 & 107 (64.1\%) & 60 (35.9\%) \\
Intent & 117 & 71 (60.7\%) & 46 (39.3\%) \\
\bottomrule
\end{tabular}
\\[6pt]\textbf{(b) Fewer-waypoint subset (50\% of questions)}\\[1pt]
\begin{tabular}{@{}lrrr@{}}
\toprule
Task & \shortstack{First edge legal +\\endpoint correct} & \shortstack{Illegal\\later edge} & \shortstack{Complete\\route legal} \\
\midrule
Point & 583 & 229 (39.3\%) & 354 (60.7\%) \\
Embodied & 104 & 68 (65.4\%) & 36 (34.6\%) \\
Intent & 72 & 43 (59.7\%) & 29 (40.3\%) \\
\bottomrule
\end{tabular}
\captionof{table}{Route outcomes after the first step and endpoint are correct, pooled over ten models. Each row divides these predictions into routes with an illegal later edge and routes whose complete edge sequence is legal. Panel (b) repeats the diagnostic on the half of questions with fewer displayed waypoints.}
\label{tab:route_construct_bridge}
\end{center}

\paragraph{Effect of reference length.}
Table~\ref{tab:route_length_stratification} stratifies outcomes by the number of reference edges. Illegal-edge rates on one-edge references are 86.3\% for Embodied and 85.5\% for Intent, rising to 92.4\% and 90.2\% for references with at least three edges.

\begin{center}
\small
\setlength{\tabcolsep}{4pt}
\begin{tabular}{@{}llrrrr@{}}
\toprule
Task & Edges & $N$ & End hit & Success & Illegal edge \\
\midrule
Point & 1 & 250 & 31.9 & 20.4 & 44.8 \\
Point & $\geq$2 & 59 & 29.3 & 14.2 & 58.0 \\
\midrule
Embodied & 1 & 135 & 5.7 & 2.7 & 86.3 \\
Embodied & 2 & 90 & 6.1 & 1.4 & 88.9 \\
Embodied & $\geq$3 & 84 & 6.0 & 1.2 & 92.4 \\
\midrule
Intent & 1 & 91 & 6.4 & 3.2 & 85.5 \\
Intent & 2 & 56 & 9.1 & 2.7 & 87.3 \\
Intent & $\geq$3 & 54 & 3.3 & 0.4 & 90.2 \\
\bottomrule
\end{tabular}
\captionof{table}{Route results by reference length, pooled over ten VLMs (\%). $N$ counts questions before model expansion.}
\label{tab:route_length_stratification}
\end{center}

\section{Human--Model Comparisons and Qualitative Examples}
\label{app:human_examples}

\subsection{Same-Question Human--Model Comparison}
We compare human responses with ten models on 50 fixed questions, with ten per task. The human reference scores 54.2 and GPT-6 Astra, the strongest VLM on this subset, scores 48.7. Averaged over the three route tasks, the human reference achieves 70.0\% valid paths and 46.7\% success, compared with 66.7\% and 33.3\% for Astra.

\begin{center}
\begin{minipage}{\columnwidth}
\centering
\small
\setlength{\tabcolsep}{5pt}
\begin{tabular}{@{}lrr@{}}
\toprule
Evaluator & Score & Valid path \\
\midrule
Human reference & \textbf{54.2} & \textbf{70.0} \\
GPT-6 Astra & 48.7 & 66.7 \\
GPT-5.5 & 28.6 & 26.7 \\
Claude Opus 4.8 & 26.6 & 23.3 \\
Gemini 3.1 Pro & 23.2 & 26.7 \\
Llama 4 Maverick & 18.9 & 20.0 \\
MiniMax M3 & 17.6 & 16.7 \\
Qwen 3.6 & 15.6 & 20.0 \\
Mistral Large 3 & 11.3 & 16.7 \\
Kimi K2.6 & 9.7 & 13.3 \\
Grok 4.3 Fast & 0.8 & 10.0 \\
\bottomrule
\end{tabular}
\captionof{table}{Same-question comparison on a fixed 50-question subset (10 per task; \%). Score combines chance-adjusted traversability skill and route success; valid path is averaged over route tasks.}
\label{tab:human_calibration}
\end{minipage}
\end{center}

\subsection{Qualitative Examples}

Figures S1--S10 present two real questions from each EgoPathBench task. Every panel within a figure repeats the identical first-person image and visual candidate interface, then overlays the recorded answer from the human reference or one of the ten evaluated VLMs. The overlays are generated from the returned waypoint IDs and the released evaluator records; they were not visible to respondents.

\clearpage
\onecolumn
\begin{center}
\includegraphics[width=0.96\textwidth]{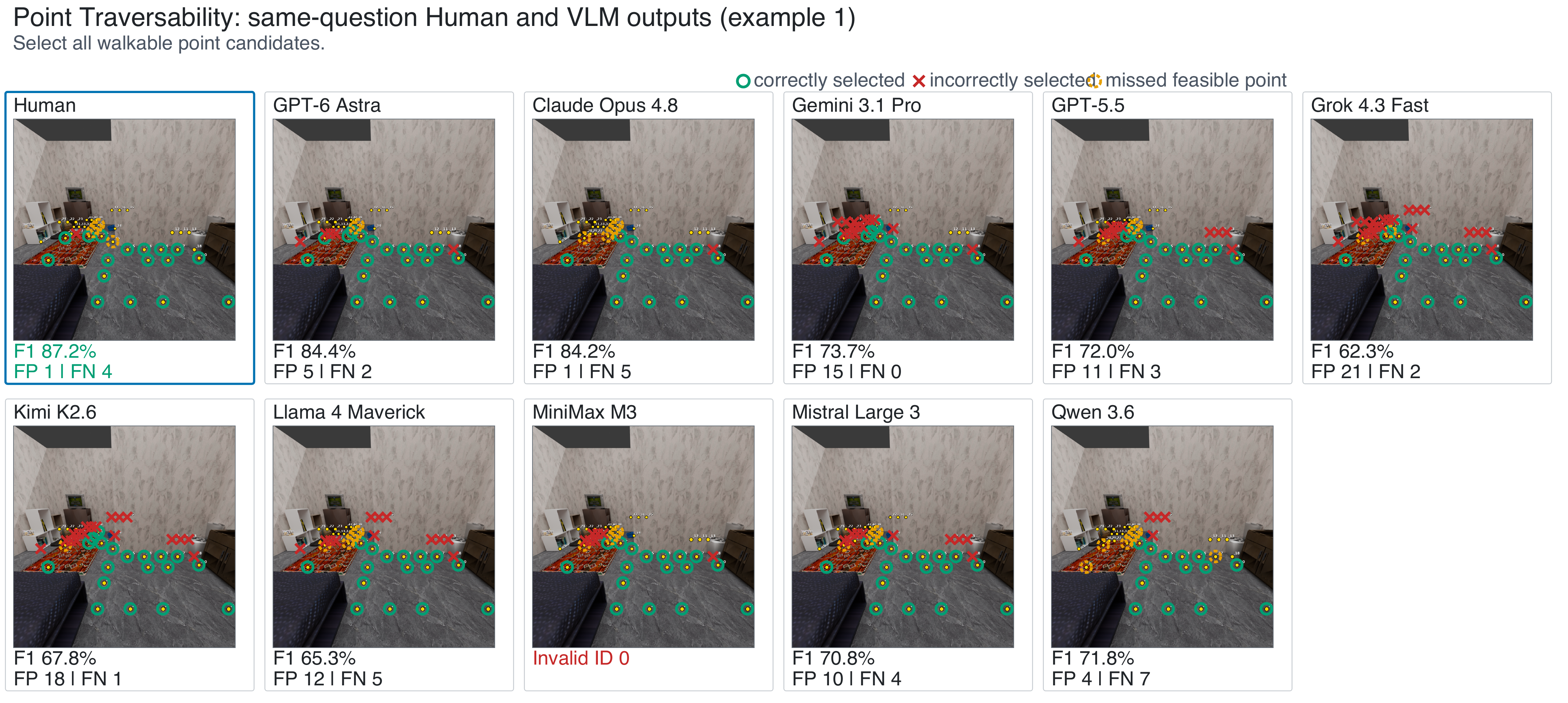}
\captionof{figure}{Same-question outputs for Point Traversability. Each panel repeats the identical first-person input and overlays the actual human-reference or model response. Green rings denote correctly selected feasible points, red crosses denote incorrectly selected points, and dashed orange rings denote feasible points omitted from the returned set.}
\label{fig:supp_a1_1_responses}
\end{center}

\begin{center}
\includegraphics[width=0.96\textwidth]{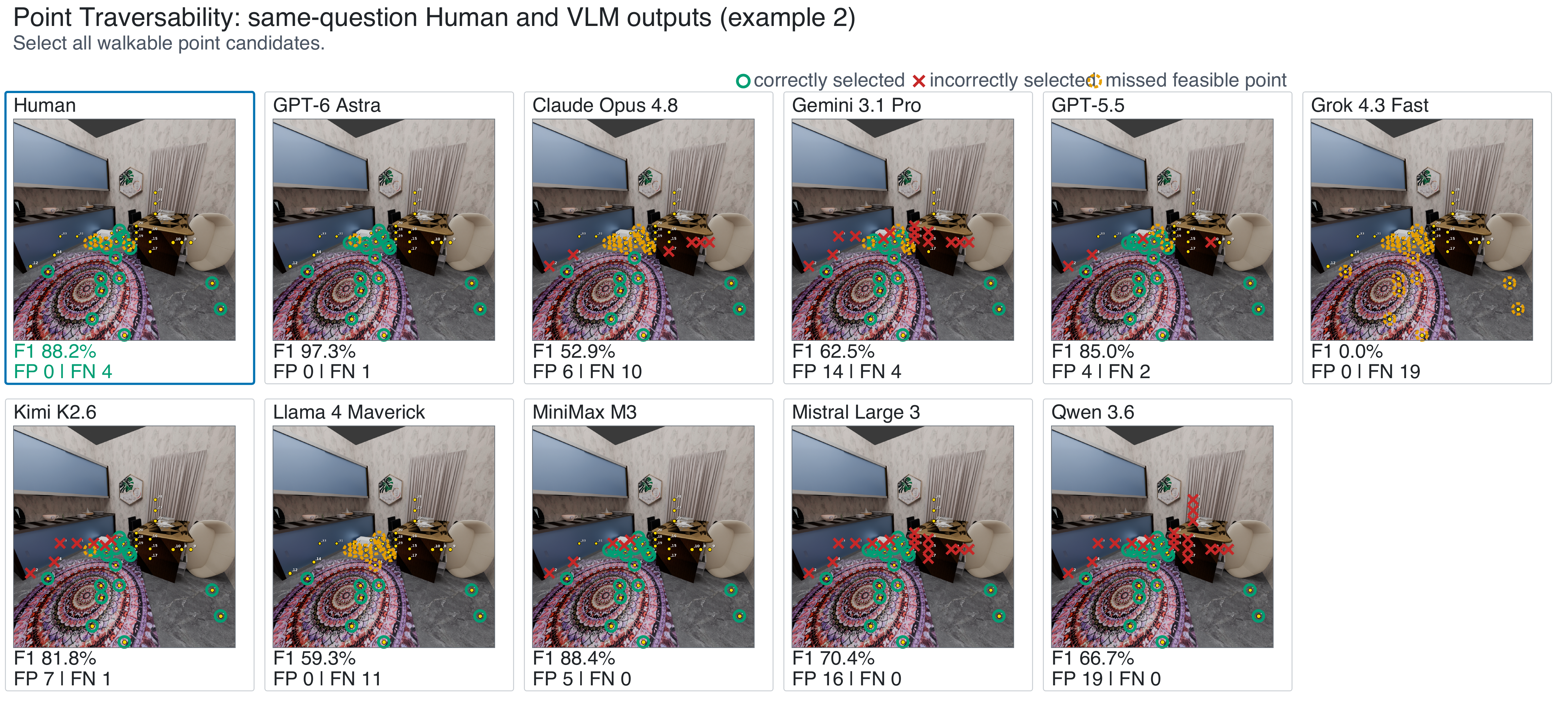}
\captionof{figure}{Same-question outputs for Point Traversability. Each panel repeats the identical first-person input and overlays the actual human-reference or model response. Green rings denote correctly selected feasible points, red crosses denote incorrectly selected points, and dashed orange rings denote feasible points omitted from the returned set.}
\label{fig:supp_a1_2_responses}
\end{center}

\begin{center}
\includegraphics[width=0.96\textwidth]{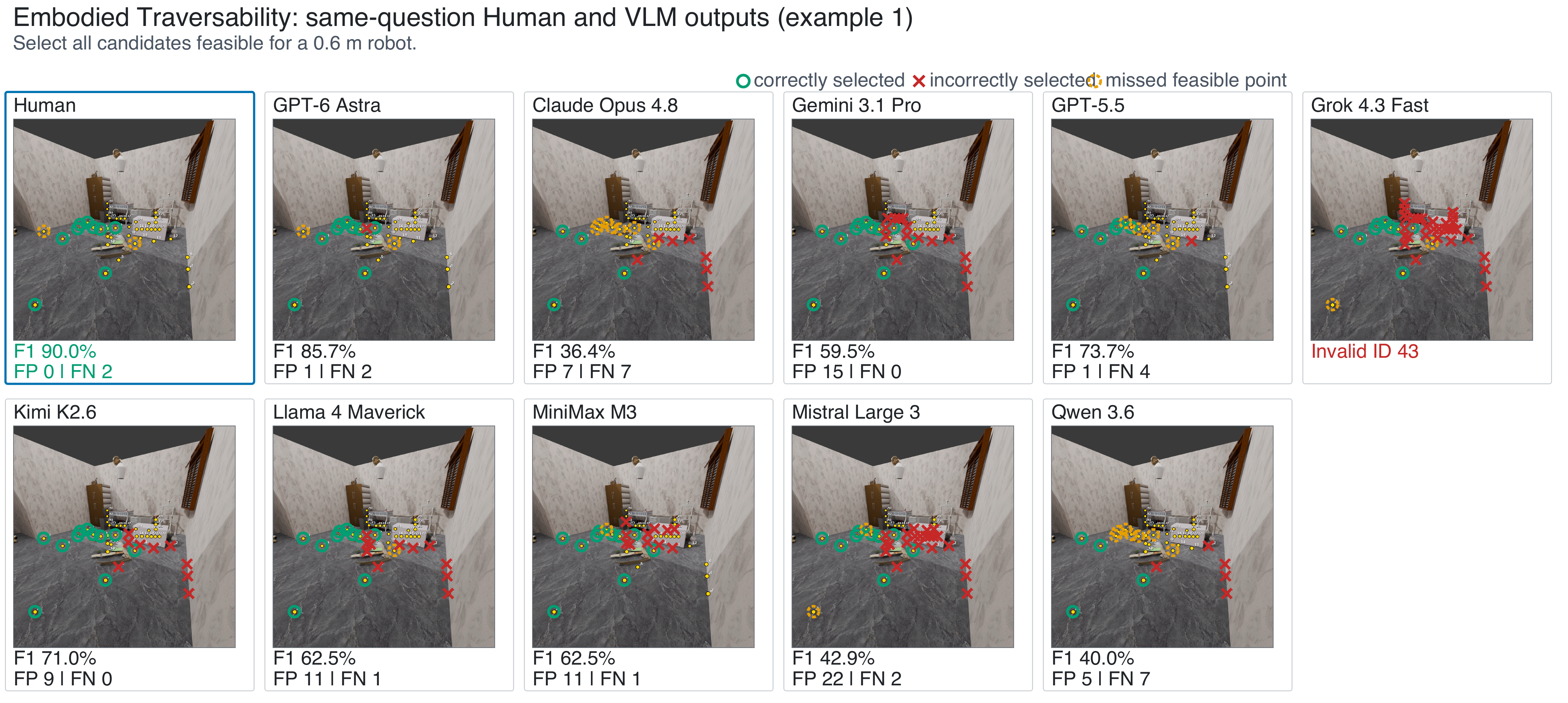}
\captionof{figure}{Same-question outputs for Embodied Traversability. Each panel repeats the identical first-person input and overlays the actual human-reference or model response. Green rings denote correctly selected feasible points, red crosses denote incorrectly selected points, and dashed orange rings denote feasible points omitted from the returned set.}
\label{fig:supp_b1_1_responses}
\end{center}

\begin{center}
\includegraphics[width=0.96\textwidth]{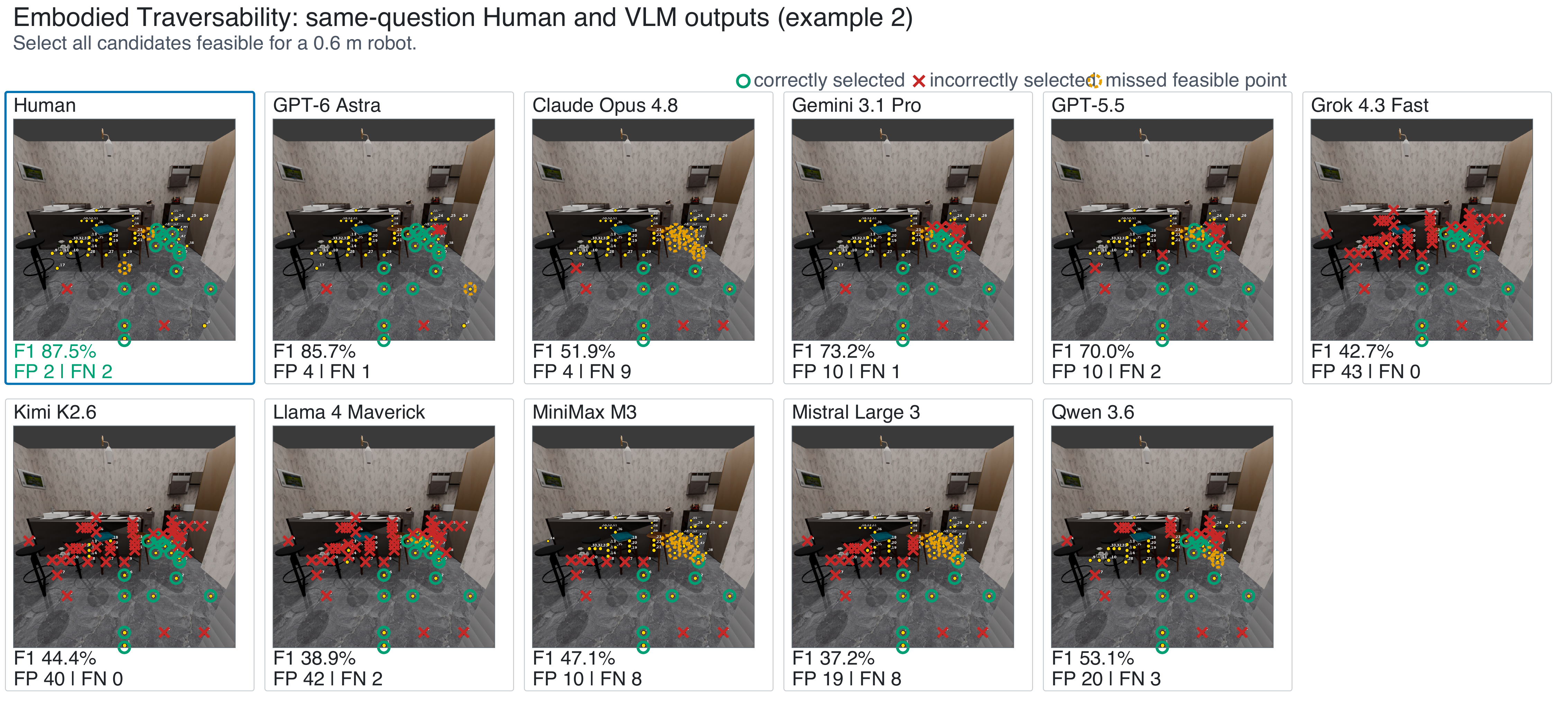}
\captionof{figure}{Same-question outputs for Embodied Traversability. Each panel repeats the identical first-person input and overlays the actual human-reference or model response. Green rings denote correctly selected feasible points, red crosses denote incorrectly selected points, and dashed orange rings denote feasible points omitted from the returned set.}
\label{fig:supp_b1_2_responses}
\end{center}

\begin{center}
\includegraphics[width=0.96\textwidth]{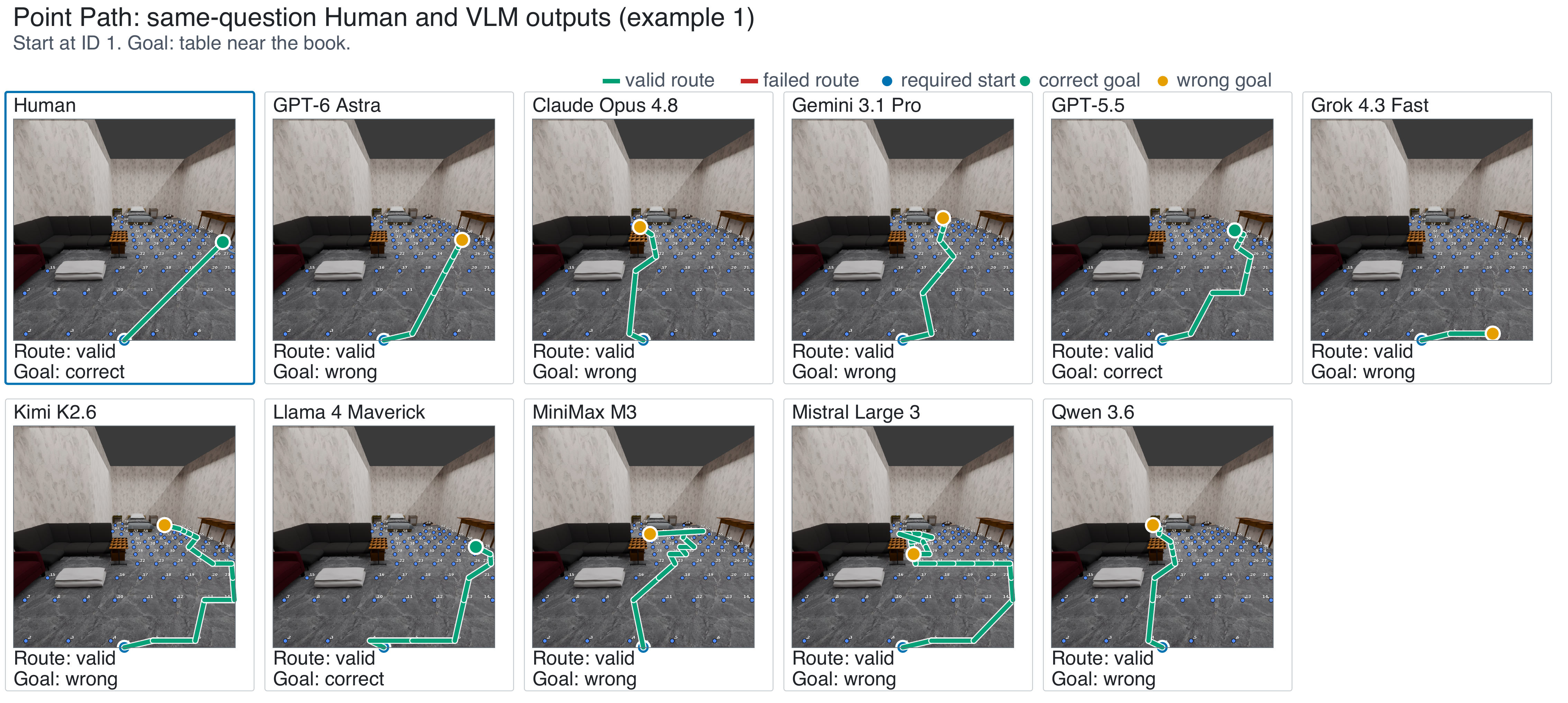}
\captionof{figure}{Same-question outputs for Point Path. Each panel repeats the identical first-person input and overlays the actual human-reference or model response. The complete returned route is green when all consecutive edges are legal and red otherwise. The required start is blue; the returned endpoint is green for an acceptable goal and orange for a wrong goal.}
\label{fig:supp_a2_1_responses}
\end{center}

\begin{center}
\includegraphics[width=0.96\textwidth]{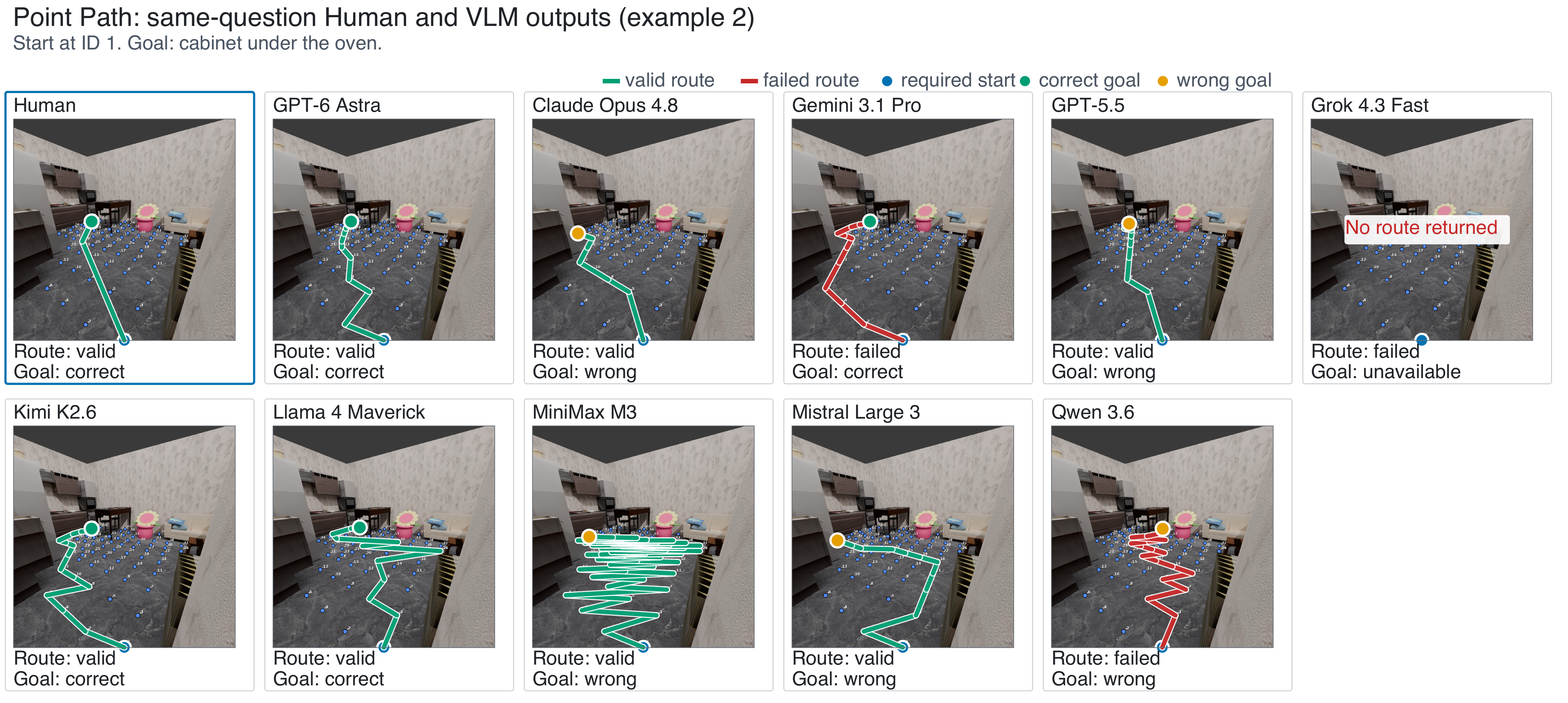}
\captionof{figure}{Same-question outputs for Point Path. Each panel repeats the identical first-person input and overlays the actual human-reference or model response. The complete returned route is green when all consecutive edges are legal and red otherwise. The required start is blue; the returned endpoint is green for an acceptable goal and orange for a wrong goal.}
\label{fig:supp_a2_2_responses}
\end{center}

\begin{center}
\includegraphics[width=0.96\textwidth]{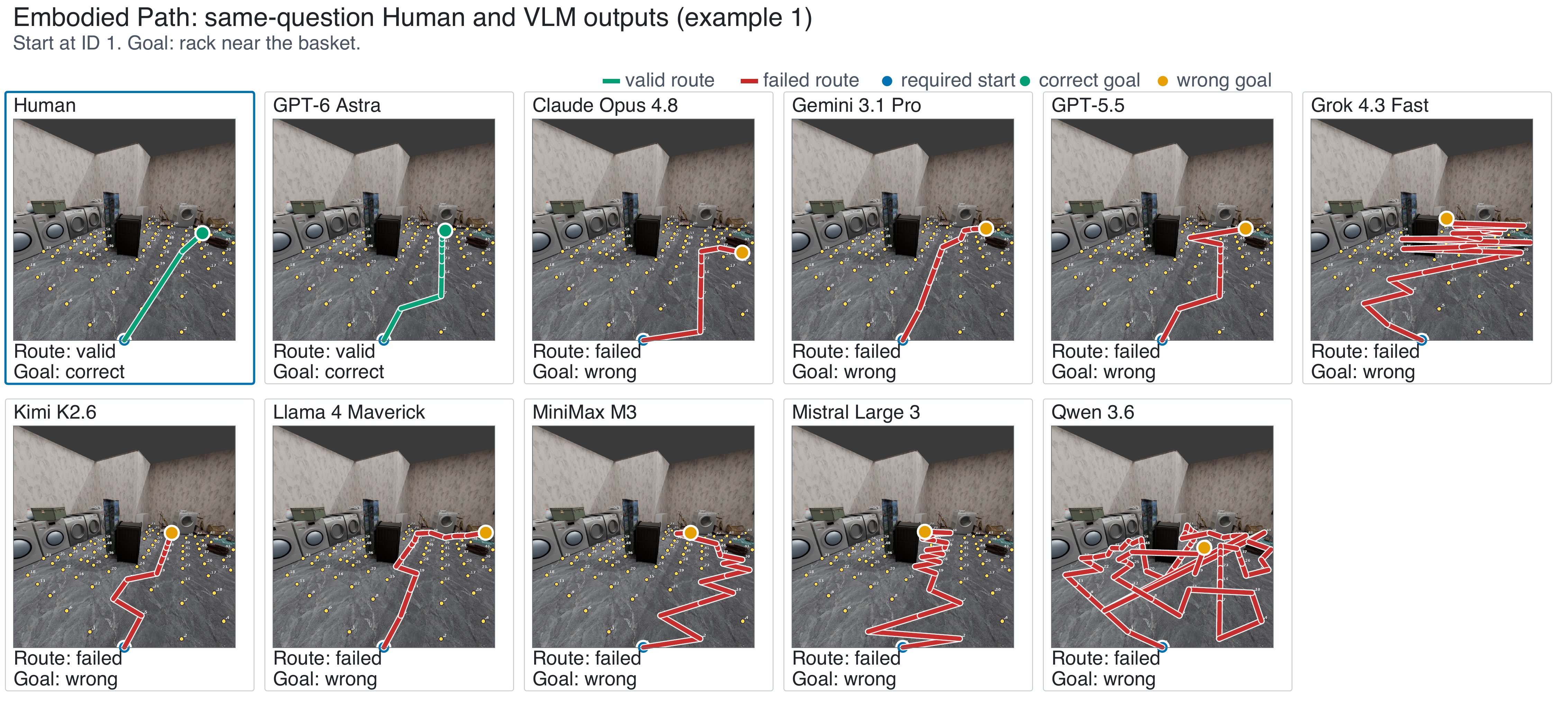}
\captionof{figure}{Same-question outputs for Embodied Path. Each panel repeats the identical first-person input and overlays the actual human-reference or model response. The complete returned route is green when all consecutive edges are legal and red otherwise. The required start is blue; the returned endpoint is green for an acceptable goal and orange for a wrong goal.}
\label{fig:supp_b2_1_responses}
\end{center}

\begin{center}
\includegraphics[width=0.96\textwidth]{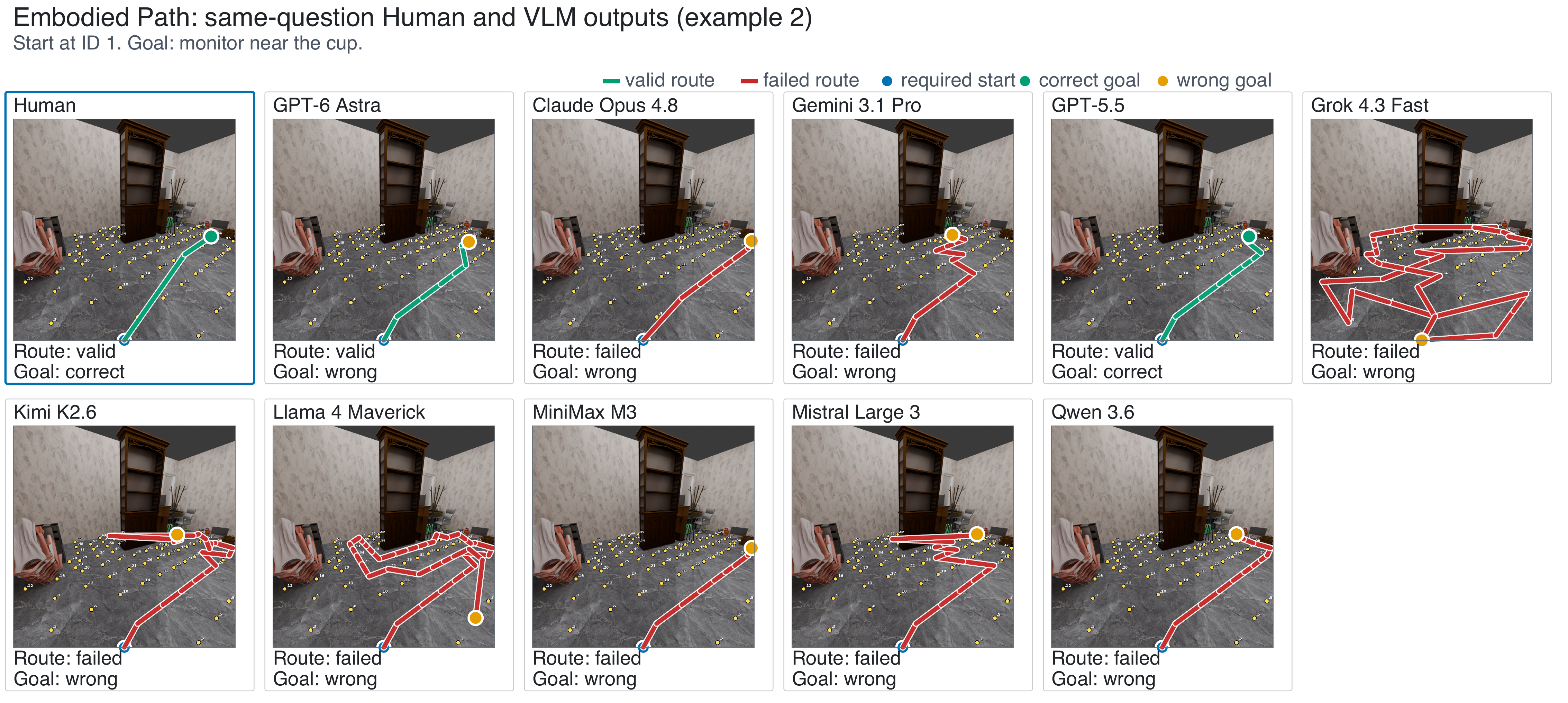}
\captionof{figure}{Same-question outputs for Embodied Path. Each panel repeats the identical first-person input and overlays the actual human-reference or model response. The complete returned route is green when all consecutive edges are legal and red otherwise. The required start is blue; the returned endpoint is green for an acceptable goal and orange for a wrong goal.}
\label{fig:supp_b2_2_responses}
\end{center}

\begin{center}
\includegraphics[width=0.96\textwidth]{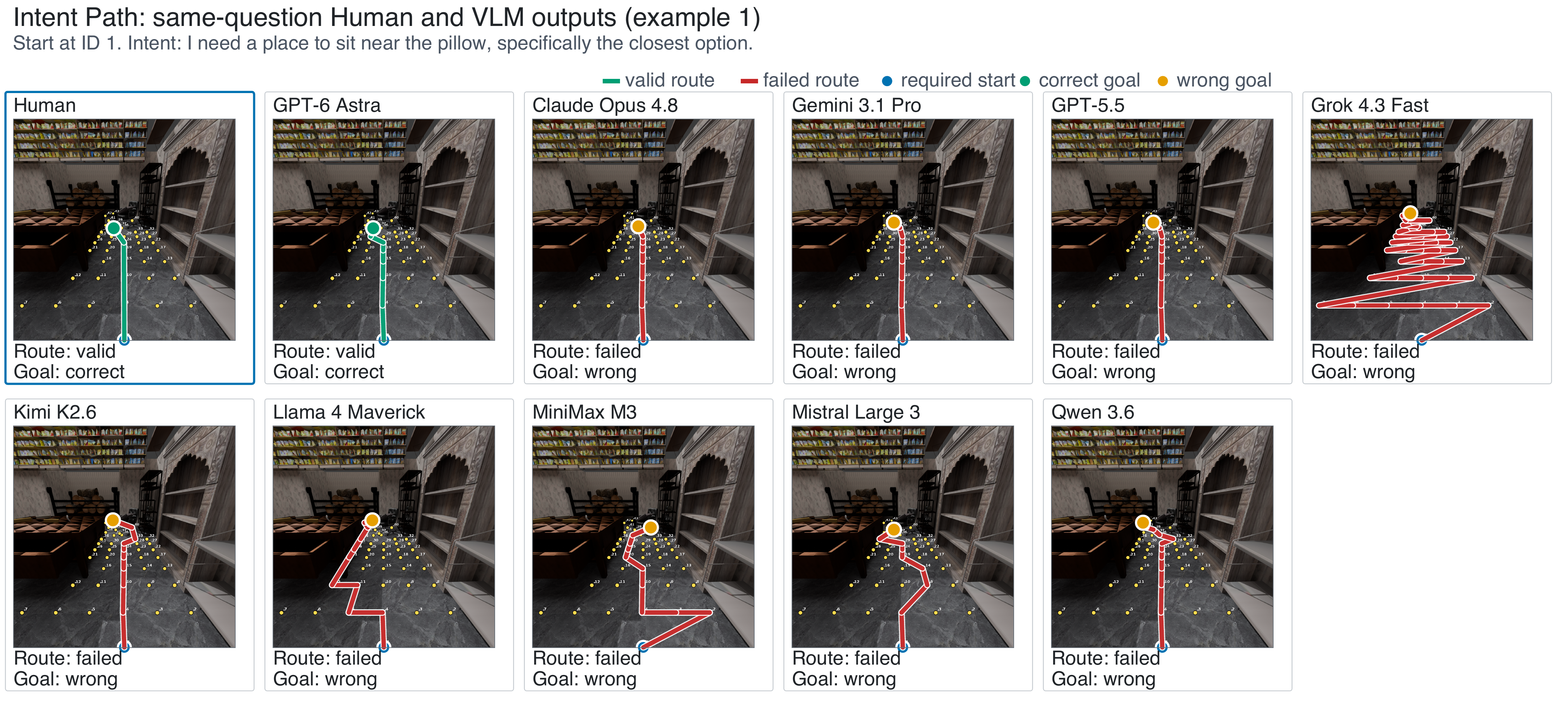}
\captionof{figure}{Same-question outputs for Intent Path. Each panel repeats the identical first-person input and overlays the actual human-reference or model response. The complete returned route is green when all consecutive edges are legal and red otherwise. The required start is blue; the returned endpoint is green for an acceptable goal and orange for a wrong goal.}
\label{fig:supp_c_1_responses}
\end{center}

\begin{center}
\includegraphics[width=0.96\textwidth]{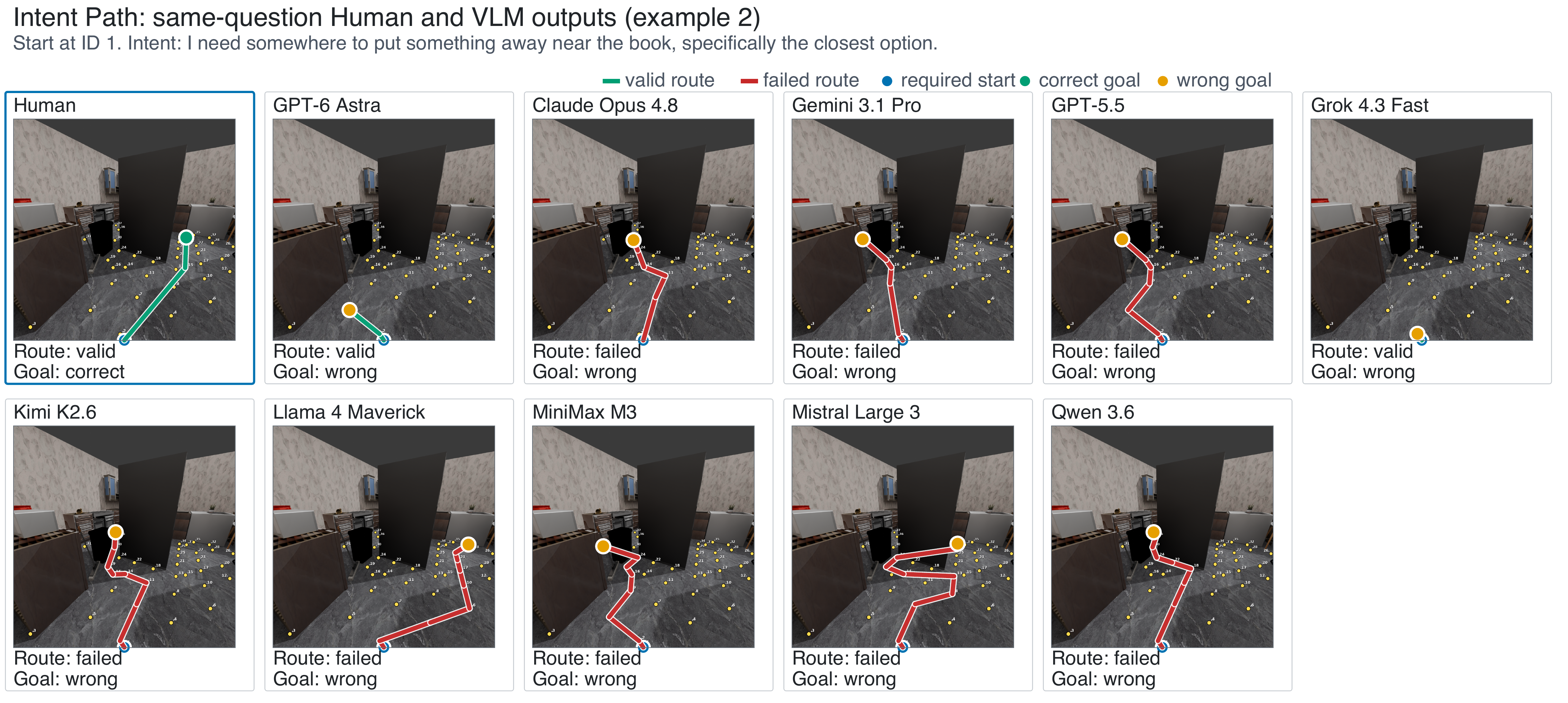}
\captionof{figure}{Same-question outputs for Intent Path. Each panel repeats the identical first-person input and overlays the actual human-reference or model response. The complete returned route is green when all consecutive edges are legal and red otherwise. The required start is blue; the returned endpoint is green for an acceptable goal and orange for a wrong goal.}
\label{fig:supp_c_2_responses}
\end{center}

\end{document}